\documentclass[journal,twoside,web]{ieeecolor}
\usepackage{jsen}
\usepackage{cite}
\usepackage{amsmath,amssymb,amsfonts}
\usepackage{graphicx}
\usepackage{textcomp}
\usepackage{wrapfig}
\usepackage{tabularx,booktabs}
\usepackage{lipsum}
\usepackage{xfrac}
\usepackage{subfig}
\usepackage{mathtools}
\usepackage{mathtools}
\DeclarePairedDelimiter\abs{\lvert}{\rvert}%

\DeclarePairedDelimiter\floor{\lfloor}{\rfloor}
\newcolumntype{b}{X}
\newcommand\scalemath[2]{\scalebox{#1}{\mbox{\ensuremath{\displaystyle #2}}}}
\newcolumntype{C}{>{\centering\arraybackslash}X} 
\newcolumntype{s}{>{\hsize=.3 \hsize}X}

\usepackage{algorithm,algpseudocode}
\algnewcommand{\algorithmicforeach}{\textbf{for each}}
\algdef{SE}[FOR]{ForEach}{EndForEach}[1]
  {\algorithmicforeach\ #1\ \algorithmicdo}
  {\algorithmicend\ \algorithmicforeach}

\usepackage{multirow}
\def\BibTeX{{\rm B\kern-.05em{\sc i\kern-.025em b}\kern-.08em
    T\kern-.1667em\lower.7ex\hbox{E}\kern-.125emX}}
\definecolor{abstractbg}{rgb}{0.89804,0.94510,0.83137}
\begin{document}
\title{Orthogonal Features Based EEG Signals Denoising Using Fractional and Compressed One-Dimensional CNN AutoEncoder}
\author{Subham Nagar and Ahlad Kumar
\thanks{Subham Nagar is currently pursuing his Master's degree in Information and Communication Technology from DA-IICT, Gandhinagar, Gujarat (e-mail: subhamnagar@gmail.com)}
\thanks{Dr. Ahlad Kumar is an Assistant Professor  at DA-IICT. (e-mail: ahlad\_kumar@daiict.ac.in). URL:https://www.daiict.ac.in/profile/ahlad-kumar/}
}

\IEEEtitleabstractindextext{%
\fcolorbox{abstractbg}{abstractbg}{%
\begin{minipage}{\textwidth}%
\begin{wrapfigure}[10]{r}{3in}%
\includegraphics[width=3in]{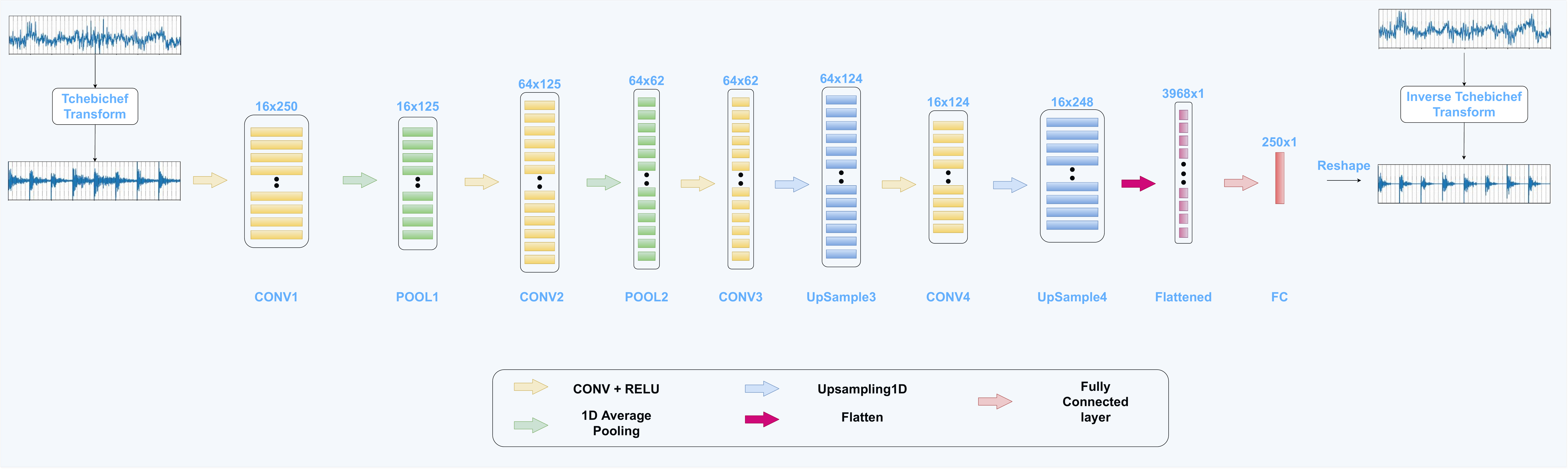}%
\end{wrapfigure}
\begin{abstract}
This paper presents a fractional one-dimensional convolutional neural network (CNN) autoencoder for denoising the Electroencephalogram (EEG) signals which often get contaminated with noise during the recording process, mostly due to muscle artifacts (MA), introduced by the movement of muscles. The existing EEG denoising methods make use of decomposition, thresholding and filtering techniques. In the proposed approach, EEG signals are first transformed to orthogonal domain using Tchebichef moments before feeding to the proposed architecture. A new hyper-parameter ($\alpha$) is introduced which refers to the fractional order with respect to which gradients are calculated during back-propagation. It is observed that by tuning $\alpha$, the quality of the restored signal improves significantly. Motivated by the high usage of portable low energy devices which make use of compressed deep learning architectures, the trainable parameters of the proposed architecture are compressed using randomized singular value decomposition (RSVD) algorithm. The experiments are performed on the standard EEG datasets, namely, Mendeley and Bonn. The study shows that the proposed fractional and compressed  architecture performs better than existing state-of-the-art signal denoising methods.
\end{abstract}

\end{minipage}}}

\maketitle

\section{Introduction}
\label{sec:introduction}
\IEEEPARstart{E}{lectroencephalogram}(EEG) is the recording of electrical activity inside the human brain and it is recorded using electrodes which are attached to the human scalp \cite{b1},\cite{b2}. During the recording process of EEG signals, they often get contaminated with various types of artifacts, due to muscle activity, eye movements and heart rhythms, which are measured by electromyogram (EMG), electrooculogram (EOG) electrocardiogram (ECG) signals, respectively \cite{b3}. Among these, the Electromyogram/muscle artifact (EMG/MA) is one such type of noise that is generally found to be challenging to eliminate, mainly due to its high amplitude and its broad frequency and anatomical distributions \cite{b4}.

Different approaches have been reported in the existing literature to remove muscle artifacts (MA) from the contaminated EEG signals. Adaptive filters \cite{b5}, low-pass filters \cite{b6}, and filter banks \cite{b7} are employed for solving the problem of signal denoising. Novel methods like RLS\cite{b8}, LMS \cite{b9}, and Kalman Filter \cite{b10} have been proposed. Numerous decomposition techniques like wavelet transform\cite{b11}, empirical mode decomposition (EMD)\cite{b12}, ensemble empirical mode decomposition (EEMD)\cite{b13} have also been employed to achieve good results. Canonical correlation analysis has also been successfully used with the above decomposition techniques\cite{b14,b15}. Sweeny et al. \cite{b14} has used both EMD and EEMD with canonical correlation analysis. Recently, variational mode decomposition (VMD) was proposed \cite{b16}, to yield superior results.

The application of machine learning and deep learning architectures have also been found to be very effective in denoising signals \cite{b17}. Denoising auto-encoder is one such kind of deep learning architecture that has outperformed existing non-deep learning based denoising methods \cite{b18},\cite{b19},\cite{b20}. However, the deep learning approaches do not address the effect of compression on signal denoising as the number of trainable weights used in the architecture increases. This can cause redundancy in the weights used and memory issues, when deployed on low energy devices. In recent years, researchers have proposed several techniques to combat redundancy in the neural network weights. Thresholding techniques like pruning is proposed in \cite{b21} to remove the least important trainable weights from the neural network. Another method used for compression calculates the low-rank approximation of weight matrices that simultaneously reduced storage and time complexity during the training and testing phases \cite{b22},\cite{b23}. The concept of randomized singular valued decomposition (RSVD) was introduced in \cite{b34}, which represented a faster way of calculating low-rank approximations as compared to the singular valued decomposition (SVD). This idea was also explored in \cite{b35} for working with large-scale data and was found to be quite effective. In this paper, the RSVD algorithm is used for compressing the trainable weight matrices of the proposed architecture.

Deep learning methods have also been employed with input features transformed to frequency domain for achieving high speed deep learning architectures\cite{b24}. The discrete cosine transform (DCT) coefficients of the input images are used to represent its important features so that neural networks can learn the image manifold in a better way and yield superior image denoising results \cite{b25}. Recently, orthogonal moment domain, has also been recently explored to address common problems in image processing \cite{b27,b28}. One such kind of orthogonal moments, namely, Tchebichef moments (TM) exhibit an essential property of energy compaction, that led to promising results in denoising images \cite{b29}. This research finding has motivated us to exploit the advantage of feeding these TM based orthogonal features to the proposed one-dimensional convolution neural network (CNN) architecture.

Traditional CNN architectures uses integer order calculus for calculating gradients during backpropagation. With fractional calculus now getting popular for solving significant problems in the image processing domain like image denoising \cite{b30,b31} and texture enhancement \cite{b32}. It has also found a place in the neural networks where gradients are calculated using differentiation with respect to a particular fractional order ($\alpha$) and has given a performance boost in classification problems as compared to conventional neural networks \cite{b33}. We have also used fractional calculus in the back-propagation phase of the proposed architecture for improved performance in the case of EEG signal denoising.

This paper is organized as follows. Section II provides some mathematical preliminaries that are useful in understanding the underlying mechanics of our architecture. Section III and IV propose the workflow of our fractional one-dimensional CNN and it's compressed form respectively. Experimental results are provided in Section V that contains the details of the datasets used, data preparation for training, performance metrics and evaluation of our proposed model under compression followed by discussion on the results obtained. Section VI concludes this work.

\section{Mathematical Preliminaries}

\begin{figure*}
\centering
 \includegraphics[width=\textwidth,height=3in]{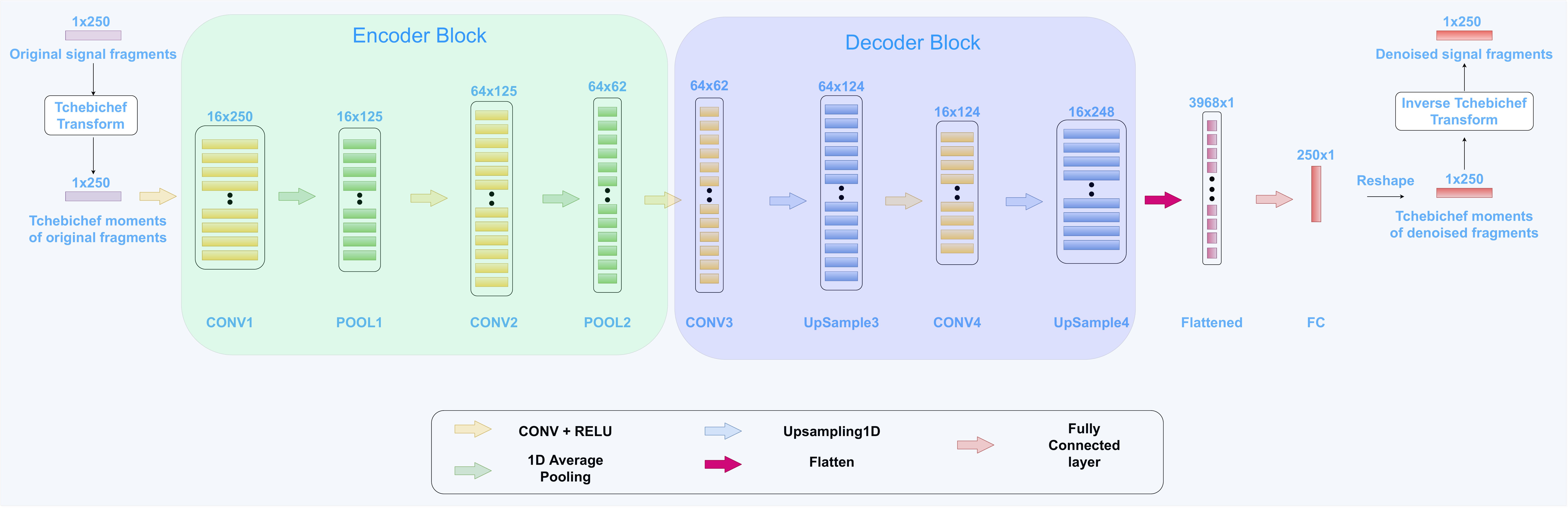}
\caption{Architecture of the proposed fractional CNN auto-encoder}
\label{cnn_fg}
\end{figure*}

\subsection{Tchebichef Moments (TM)}

Let $x(n)$ be an EEG signal with $n=1,2, \ldots, \mathcal{N}$. The relationship between the noisy signal $y(n)$ and the original signal $x(n)$ corrupted by noise is given as follows:
\begin{equation}
   y(n) = x(n)+ \zeta(n)
\label{tche_1}
\end{equation}
where $\zeta(n)$ is the  muscle  artifacts (MA) noise. This paper proposes a deep learning architecture which recovers an estimate of the original signal from its noisy observation $y(n)$. 

The Tchebichef moments of order $p$ for a signal $x(n)$ of length $\mathcal{N}$ samples is given by \cite{b26}:
\begin{equation}
    T_{p}(x) = \sum_{x=0}^{\mathcal{N}-1} t_{p}(x;\mathcal{N}) x(n)
\label{gh}
\end{equation}
with $p = 0, 1 , 2 ..... \mathcal{N}-1$. For simplicity, $t_{p}(x)$ has been used to represent $t_{p}(x;\mathcal{N})$ which is the orthonormal Tchebichef polynomials given by 
\begin{equation}
    t_{p}(x) = \beta\textsubscript{1} (2x+1-\mathcal{N}) t_{p-1}(x) + \beta\textsubscript{2} t_{p-2}(x)
\label{}
\end{equation}
where
\begin{equation}
    \beta\textsubscript{1} = \frac{1}{n}\sqrt{\frac{4(n^{2}-1)}{\mathcal{N}^{2}-n^{2}}}
\label{}
\end{equation}
\begin{equation}
    \beta\textsubscript{2} = \frac{1-n}{n}\sqrt{\frac{2n+1}{2n+3}}\sqrt{\frac{\mathcal{N}^{2}-(n-1)^{2}}{\mathcal{N}^{2}-n^{2}}}
\label{}
\end{equation}
The initial conditions for the recurrence relations are 
\begin{equation}
    t\textsubscript{0}(x) = \frac{1}{\sqrt{n}}
\label{}
\end{equation}
and
\begin{equation}
    t\textsubscript{1}(x) = (2x+1-\mathcal{N})\sqrt{\frac{3}{\mathcal{N}(\mathcal{N}^{2}-1)}}
\label{}
\end{equation}
The set of TMs upto order $p$ in matrix form is given as
\begin{equation}
    T_{p}(\textbf{X}) = \textbf{X}\textbf{Q}^{T}
\label{c1p}
\end{equation}
where $\textbf{X} = [x(0),\; x(1),\; x(2), \ldots,\; x(\mathcal{N}-1)]$ and 
\begin{equation}
\textbf{Q} = \begin{bmatrix} 
    t\textsubscript{0}(0) & \dots  & t\textsubscript{0}(\mathcal{N}-1) \\
    \vdots & \ddots & \vdots\\
    t_{p-1}(0) & \dots  & t_{p-1}(\mathcal{N}-1)
    \end{bmatrix}
\label{}
\end{equation}
Here, $\textbf{Q}$ is the Tchebichef polynomial matrix upto order $p$. The original one-dimensional signal $\textbf{X}$ can be reconstructed  from the set of Tchebichef moments using the following equation
\begin{equation}
    \textbf{X} = T_{p}(\textbf{X})\textbf{Q}
\label{c10}
\end{equation}

\subsection{Compression using RSVD}
The compression of the kernel and weight matrices used in the proposed architecture is carried out using low rank approximation of these matrices. It is done using the RSVD technique, which decomposes the original matrix $ A \in  \mathbb{R}^{n\times m} $ into a smaller randomized subspace $B \in  \mathbb{R}^{c \times m}$, where $c < n$. For kernel matrix $ K \in  \mathbb{R}^{n\times c\times f} $, where $n$, $c$ and $f$ refer to the number of filters, number of channels and feature dimension respectively, we reshape it into a $2D$ matrix $A$ of the form $\mathbb{R}^{n\times m}$, where $m=c*f$.

For calculating the low rank approximation, let $O \in  \mathbb{R} ^{m \times (r+p)}$ be a normally distributed random matrix, where $r$ is the rank to be approximated, $p$ denotes the number of additional projections such that $r+p < n$. We define $Q\textsubscript{i}$ as the orthogonal basis after $i$ iterations, where $ i = 1,2,3 \ldots, k$ and $k$ denotes the number of subspace iterations. The value of $Q\textsubscript{0}$ is set using the following equation
\begin{equation}
    Q\textsubscript{0} = AO
\label{}
\end{equation}
The recurrence relation for calculating the orthogonal basis $Q\textsubscript{i}$ is given by
\begin{equation}
   G\textsubscript{i} = qr(A\textsuperscript{T}Q\textsubscript{i-1})
\label{c1}
\end{equation}
\begin{equation}
Q\textsubscript{i}  = qr(AG\textsubscript{i})
\label{}
\end{equation}
where $qr()$ is the function for the QR decomposition operation which factorizes a matrix into an orthogonal matrix and an upper triangular matrix. Here, we just take the orthogonal matrix part. The above recurrence equations are applied over $k$ iterations, after which, we get the value $Q_k$. The condensed matrix $B$ can be calculated as
\begin{equation}
   B=Q\textsubscript{k}\textsuperscript{T}A
\label{}
\end{equation}
The SVD decomposition of this condensed matrix is
\begin{equation}
   [U_{r} , S_{r}, V_{r}] = \text{SVD}(B)
\label{}
\end{equation}
\begin{equation}
U\textsubscript{r} = Q\textsubscript{k}U\textsubscript{r}
\label{c1}
\end{equation}
where, $U_{r}\in \mathbb{R} ^ {n\times r} $, $V_{r}\in \mathbb{R} ^{m\times r}$ are the matrices with orthonormal columns and $S_{r}\in \mathbb{R}^{r\times r}$  is a diagonal matrix.

\subsection{Fractional Order Processing}
Unlike the integer order derivatives, various definitions have been proposed for fractional order derivatives. The three most commonly used fractional order derivatives are, namely, Grunwald Letnikov (G-L), Riemann-Liouville (R-L) and Caputo derivatives \cite{b33}. We have used Caputo fractional derivative (CFD) of a function $f(x)$ with order $\alpha$, defined as follows:
\begin{equation}
    \frac{d^{\alpha}f(x)}{dx^{\alpha}}= \frac{1}{\Gamma(n-\alpha)}
    \int_{a}^x \frac{f^{(n)}(y)}{(x-y)^{(1+\alpha-n)}}dy
\label{frac_eq1}
\end{equation} 
where $n-1 < \alpha < n$ , $n \in \mathbb{N^{+}}$, $a$ is the initial value and $\Gamma(\cdot)$ denotes the Gamma function. The CFD is found to be consistent with the integer order derivatives used in neural networks, because of which this derivative is applied in several engineering problems \cite{b33}. This motivated us to employ it during the back-propagation of our proposed model. Let $\alpha$ be the fractional order for which the derivative needs to be calculated and $f(x) = (x-a)^{k}$ be a polynomial function of degree $k$. The Caputo fractional derivative is given by \cite{b36}:
\begin{equation}
\frac{d^{\alpha}f(x)}{dx^{\alpha}} = \frac{\Gamma(k+1) (x-a)^{k-\alpha}}{\Gamma(k-\alpha+1)}
\label{frac_eq2}
\end{equation}
For simplicity of the notation, the fractional derivative $\dfrac{\partial ^{\alpha}f(x)}{\partial x^{\alpha}}$ is denoted as $D_{x}^{\alpha} f(x)$ and will be used in calculating gradients of the proposed architecture.

\begin{figure*}
\centering
 \includegraphics[width=\textwidth,height=2in]{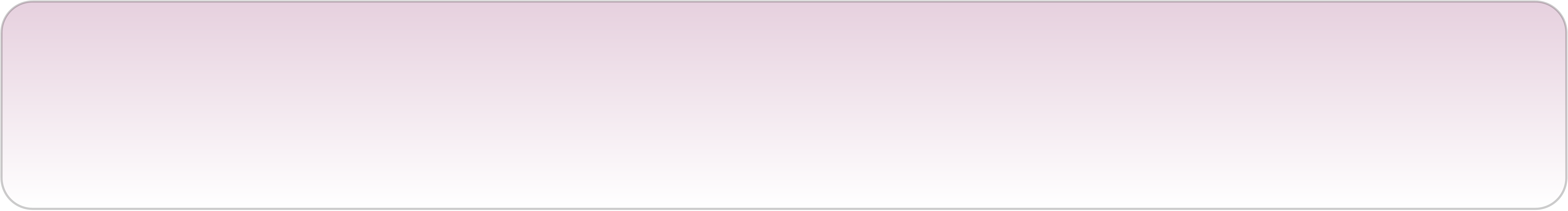}
\caption{Illustration of $im2col$ transformation on 1D input}
\label{im2col_fg}
\end{figure*}

\section{Proposed Architecture}
\label{sec_2}
The proposed fractional CNN based architecture for denoising EEG signals is shown in Fig. \ref{cnn_fg}. The model is based on the encoder-decoder architecture. The encoding of the EEG signal is carried out and the information is represented in the compressed form as latent vectors. This is followed by up-sampling (decoder) operation  that recovers the information present in the EEG signal from the latent space. This can be observed from Fig. \ref{cnn_fg}, where each of the first two convolutional layers followed by average pooling layers constitutes the encoder block while each of the last two convolutional layers followed by up-sampling layers constitutes the decoder block. Here, the up-sampling layers are used for recovering structural details present in the EEG signals. 

The original EEG signal fragments are transformed into TMs (orthogonal) space $T_{\mathcal{N}}(\textbf{X})$ using Eq. \ref{c1p} where \textbf{X} is the original signal fragment of dimension $\mathcal{N}$. Similarly, the transformation of the noisy signal $T_{\mathcal{N}}(y)$ has also been done, which is then fed as an input to the  first convolutional layer of the proposed architecture. The architecture has four convolutional layers (CONV), two average pooling layers, two upsampling layers and a flattened layer, which is connected to the fully connected (FC) layer. The first convolutional layer (CONV1) has $16$ filters. The next two convolutional layers (CONV2,$\;$CONV3) have 64 filters each while CONV4 has 16 filters.  Rectified linear units (ReLU) have been employed as activation functions for convolutional hidden layers. There are 250 neurons in the fully connected layer (FC). All the convolutional layers are having kernel of dimension $1 \times 3$ and kernel stride is taken as $1$. The average pooling layers have kernel dimension of 2 and a stride of 2, while the up-sampling layers have the up-sampling factor as $2$. The padding ensures the output dimension to be same as that of the input. Next, the workflow of the architecture that involves forward and proposed fractional backward propagation will be discussed. 

\subsection{Forward Propagation}

\subsubsection{Convolutional Layer}
The input-output relationship for the convolutional layer of the architecture is given as
\begin{equation}
    S_{m,j}^{[i]} = \sum_{n=0}^{N_C^{[i-1]}-1} \sum_{p=0}^{F^{[i]}-1} I_{n,j+p}^{[i-1]}K_{m,n,p}^{[i]} + b_m^{[i]}
\label{convfrwd}
\end{equation}
where $m=0...N_F^{[i]}-1$, $j=0...N_{W}^{[i]}-1$.
Here, $I^{[i-1]}$ is the input of dimension  $N_C^{[i-1]} \times N_{W}^{[i-1]}$, being fed to the $i^{th}$ convolutional layer with $N_C^{[i-1]}$ representing the number of channels and $N_{W}^{[i-1]}$ being the feature dimension. For the $i^{th}$ convolutional layer, we have the trainable kernel $K^{[i]}$ of size $N_F^{[i]} \times N_C^{[i-1]} \times F^{[i]}$, where $F^{[i]}$ is the kernel filter dimension and $N_F^{[i]}$ denotes the number of kernel filters.  The matrix $b^{[i]}$ denotes the bias for this layer of dimension $N_F^{[i]} \times 1$. $S^{[i]}$ is the output of the convolution layer and is of size $ N_F^{[i]} \times N_{W}^{[i]}$, where the output feature dimension $N_{W}^{[i]}$ is given by
\begin{equation}
    N_{W}^{[i]} = (N_{W}^{[i-1]}-F^{[i]}+2*g^{[i]}) + 1
\label{c1}
\end{equation}
with $g^{[i]}$, denoting the padding size.
In this paper, Tchebichef vector of the noisy signal $T_{\mathcal{N}}(y)$ is taken as the input features denoted by $I^{[0]}$ (Eq. \ref{convfrwd}). This is fed to the first  convolutional layer of the architecture, with $N_C^{[0]}=1$ and $N_W^{[0]}=\mathcal{N}$.

For faster implementation of the above approach, we use the method of matrix multiplication to represent convolution operation given in Eq. \ref{convfrwd}. For this, input $I^{[i-1]}$ is converted into matrix form using the following transformation:
\begin{equation}
   I_{col}^{[i-1]} = im2col(I^{[i-1]})
\label{conv1}
\end{equation}
where the dimension of $I_{col}^{[i-1]}$ is $(N_C^{[i-1]} \times F^{[i]} \times N_{W}^{[i]})$,
with $N_{W}^{[i]} = (N_{W}^{[i-1]}-F^{[i]}+2*g^{[i]}) + 1$

Here, $im2col$ refers to the technique in which input of size $1 \times F^{[i]}$ is taken and stack it in the form of columns of a matrix. The pictorial illustration of $im2col$ has been shown in Fig. \ref{im2col_fg}. Now, Eq. (\ref{convfrwd}) gets modified as

\begin{equation}
    S^{[i]} = K^{[i]} (*) I_{col}^{[i-1]} + b^{[i]}
\label{convv3}
\end{equation}
where (*) denotes the matrix multiplication,
$b^{[i]}$ is the bias matrix of size $(N_{F}^{[i]} \times 1)$.

Each convolutional layer output $S^{[i]}$ is fed as an input to the rectified linear unit activation function (ReLU), which gives $C^{[i]}$ as the output governed by the following expression
\begin{equation}
    C_{m,j}^{[i]} = ReLU (S_{m,j}^{[i]})
\label{relu1}
\end{equation}

\subsubsection{Average pooling Layer}
Average pooling operation is applied on the activated feature maps of the $i^{th}$ convolutional layer $C^{[i]}$ having dimension of $N_F^{[i]} \times N_{W}^{[i]}$. In this study, pooling filter size $P_f^{[i]}=2$ and stride $P_s^{[i]}=2$ is taken. The pooling output $P^{[i]}$ of dimension $N_F^{[i]} \times N_{P}^{[i]}$ is given as
\begin{equation}
    P^{[i]}_{m,k} = \frac{C_{m,2k}^{[i]}+C_{m,2k+1}^{[i]}}{2}
\label{pool1}
\end{equation}
where $m=0...N_F^{[i]}-1$ and $k=0...N_{P}^{[i]}-1$ with $N_{P}^{[i]} = \frac{N_{W}^{[i]}}{2}$.

\subsubsection{Up-sampling Layer}
Up-sampling operation increases the resolution of the activated feature maps, i.e, the $i^{th}$ convolutional layer $C^{[i]}$. The output of the up-sampling layer  $U^{[i]}$ is defined as 
\begin{equation}
 U^{[i]}_{m,j} = C^{[i]}_{m,\floor*{j/2}}
\label{upsample}
\end{equation}
where $m=0...N_F^{[i]}-1$ and $j=0...N_{U}^{[i]}-1$ with $N_{U}^{[i]} = 2*N_{W}^{[i]}$ as the up-sampling factor $U_f^{[i]}=2$. The dimension of $U^{[i]}$ is $N_F^{[i]} \times N_{U}^{[i]}$.

\subsubsection{Flattened Layer}
Before we feed the features to the FC layer, it needs to be flattened into a one-dimensional format. This process can be represented by the following equation

\begin{equation}
    \mathcal{F} = flatten(R)
\label{c1xb}
\end{equation}
where $R$ is the input for this layer and $\mathcal{F}$ is the flattened output which can be used in the FC layer.

\subsubsection{FC Layer}
Now, the forward propagation can be represented by

\begin{equation}
    \hat T_{\mathcal{N}}(x) = W (*) \mathcal{F} + B
\label{c1vb}
\end{equation}
where $\mathcal{F}$, W and B denote the output of the flattened layer, weight matrix and bias respectively. Here, $\hat T_{\mathcal{N}}(x)$ is the estimated denoised signal which will be used in the formulation of the loss function discussed next. 


\subsection{Loss function}

The proposed loss function for the fractional CNN auto-encoder is given as  

\begin{equation}
\scalemath{0.85}{
    L = \frac{1}{2\mathcal{M}} \sum_{i=0}^{\mathcal{M}-1} \left(\hat T_{\mathcal{N}}^{(i)}(x) - T_{\mathcal{N}}^{(i)}(x)\right)^{2} +\frac{\lambda}{2} \left(\sum_{j=1}^{E} \left\lVert K^{[j]} \right\rVert_{2}^{2} + \left\lVert W \right\rVert_{2}^{2}\right)
\label{reg_l2}
}
\end{equation}
where, $\mathcal{M}$ represents the number of training samples, $E$ is the number of convolutional layers and $\lambda$ represents the regularization parameter. For back-propagation, the derivative of loss with respect to $\hat T_{\mathcal{N}}(x)$ is calculated as
\begin{equation}
    \frac{\partial L}{\partial \hat T_{\mathcal{N}}(x)} = \frac{1}{\mathcal{M}} \sum_{i=0}^{\mathcal{M}-1} \left(\hat T_{\mathcal{N}}^{(i)}(x) - T_{\mathcal{N}}^{(i)}(x)\right)
\label{c1c}
\end{equation}
Eq. (\ref{c1c}) is used during back-propagation process which is discussed next.

\subsection{Fractional Back-Propagation} \label{cai}

\begin{figure*}
\centering
 \includegraphics[width=\textwidth,height=2in]{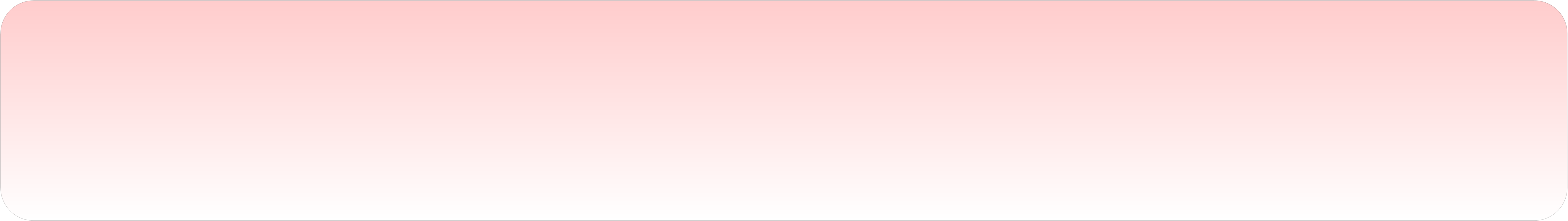}
\caption{Illustration of $col2im$ transformation}
\label{col2im_fg}
\end{figure*}
The proposed back-propagation technique comes with the advantage in the form of an extra hyper-parameter, i.e, fractional order $\alpha$ which can be tuned to obtain the best denoising performance and also plays an important role in training the architecture. 

\subsubsection{FC Layer}
The value obtained using Eq. (\ref{c1c}) is used as input to the FC layer.
Using Eqs. (\ref{c1vb}) and (\ref{reg_l2}), the fractional gradients of weights, i.e, $D_{W}^{\alpha} L$ is calculated as follows: 





\begin{equation}
    D_{W}^{\alpha} L = \left(\frac{\partial L}{\partial \hat T_{\mathcal{N}}(x)}\right)D_{W}^{\alpha} \hat T_{\mathcal{N}}(x) + \frac{\lambda}{2} D_{W}^{\alpha} {W}^2
\label{c1ah}
\end{equation}
Using Eq. (\ref{frac_eq2}), the fractional gradients present on the right hand side of Eq. (\ref{c1ah}) are calculated as follows
\begin{equation}
  D_{W}^{\alpha} \hat T_{\mathcal{N}}(x) = \mathcal{F}\frac{W^{1-\alpha} }{\Gamma (2-\alpha)}
\label{c1}
\end{equation}
\begin{equation}
  \frac{\lambda}{2} D_{W}^{\alpha} {W}^2 = \lambda \frac{W^{2-\alpha} }{\Gamma (3-\alpha)}
\label{c1}
\end{equation}
where, $\Gamma(\cdot)$ denotes the gamma function. Substituting these values obtained in Eq. (\ref{c1ah}) results in 
\begin{equation}
    D_{W}^{\alpha} L = \left(\frac{\partial L}{\partial \hat T_{\mathcal{N}}(x)}\right)\mathcal{F}\frac{W^{1-\alpha} }{\Gamma (2-\alpha)} + \lambda \frac{W^{2-\alpha} }{\Gamma (3-\alpha)}
\label{c1}
\end{equation}
Similarly, the fractional gradient with respect to the bias $B$ is given by
\begin{equation}
    D_{B}^{\alpha} L = \frac{\partial L}{\partial \hat T_{\mathcal{N}}(x)} D_{B}^{\alpha} \hat T_{\mathcal{N}}(x)   = \frac{\partial L}{\partial \hat T_{\mathcal{N}}(x)}\frac{B^{1-\alpha} }{\Gamma (2-\alpha)}
\label{c1}
\end{equation}

\subsubsection{Flattened Layer}
This layer flattens the input during forward propagation as mentioned in Eq.(\ref{c1xb}). So during backward propagation, we can reshape the output gradient $\frac{\partial L}{\partial \mathcal{F}}$ in the shape of $R^{[i]}$ to get the input gradient $\frac{\partial L}{\partial R}$

\subsubsection{Up-sampling Layer}
For backpropagating the gradients, the successive gradients from $\dfrac{\partial L}{\partial U^{[i]}}$ are added and assigned to each element in $\dfrac{\partial L}{\partial C^{[i]}}$. 
This can be represented using the following equation: 
\begin{equation}
\dfrac{\partial L}{\partial C_{m,j}^{[i]}} = \dfrac{\partial L}{\partial U_{m,2j}^{[i]}} + \dfrac{\partial L}{\partial U_{m,2j+1}^{[i]}}
\label{upsample_back}
\end{equation}
where the up-sampling factor $U_{f}^{[i]}=2$.
The dimension of $\dfrac{\partial L}{\partial C^{[i]}}$ is $N_F^{[i]} \times N_{W}^{[i]}$ where $N_{W}^{[i]}= N_{U}^{[i]} / 2$.
 The gradient $\dfrac{\partial L}{\partial C^{[i]}}$ for the $m^{th}$ feature map can also be represented in a matrix form as follows
\begin{equation}
\scalemath{0.7}{
\dfrac{\partial L}{\partial C_{m}^{[i]}}
    =     \begin{bmatrix}
    \dfrac{\partial L}{\partial U_{m,0}^{[i]}}+\dfrac{\partial L}{\partial U_{m,1}^{[i]}} & \dfrac{\partial L}{\partial U_{m,2}^{[i]}}+\dfrac{\partial L}{\partial U_{m,3}^{[i]}} & \cdots & \dfrac{\partial L}{\partial U_{m,W_u^{[i]}-2}^{[i]}} + \dfrac{\partial L}{\partial U_{m,W_u^{[i]}-1}^{[i]}}
    \end{bmatrix}
    }
\end{equation}
where, $\dfrac{\partial L}{\partial U_{m,k}^{[i]}}$ refers to the $k^{th}$ element of $m^{th}$ feature map in $\dfrac{\partial L}{\partial U^{[i]}}$.\\

\subsubsection{Average Pooling Layer}
Similar to the back-propagation for Up-sampling layer, $\frac{\partial L}{\partial C^{[i]}}$ is calculated from the pooling gradient $\frac{\partial L}{\partial P^{[i]}}$. Considering the fact that $P^{[i]}$ is the average pooling output, here the operation will be slightly different. Each element in $\frac{\partial L}{\partial P^{[i]}}$ is divided by the pooling size $P_{f}^{[i]}$ and proportionally back-propagate the error gradients to the input. 
This can be represented using the following equation: 
\begin{equation}
\dfrac{\partial L}{\partial C_{m,j}^{[i]}} = \dfrac{1}{2}\dfrac{\partial L}{\partial P_{m,\floor*{j/2}}^{[i]}}
\label{pool_back1}
\end{equation}
where, $P_{f}^{[i]}=2$. The gradient $\frac{\partial L}{\partial C^{[i]}}$ for the $m^{th}$ feature map calculated in Eq. (\ref{pool_back1}) can also be represented in a matrix form as follows

\begin{equation}
\scalemath{0.75}{
\dfrac{\partial L}{\partial C_{m}^{[i]}}
    =     \frac{1}{2}\begin{bmatrix}
    \dfrac{\partial L}{\partial P_{m,0}^{[i]}} & \dfrac{\partial L}{\partial P_{m,0}^{[i]}}  & \dfrac{\partial L}{\partial P_{m,1}^{[i]}}  &  \dfrac{\partial L}{\partial P_{m,1}^{[i]}} & \cdots
    \dfrac{\partial L}{\partial P_{m,N_{W}^{[i]}-1}^{[i]}}  &  \dfrac{\partial L}{\partial P_{m,N_{P}^{[i]}-1}^{[i]}} &
    \end{bmatrix}
    }
\end{equation}
where, $\dfrac{\partial L}{\partial P_{m,k}^{[i]}}$ refers to the $k^{th}$ element of $m^{th}$ feature map in $\dfrac{\partial L}{\partial P^{[i]}}$. The dimension of $\frac{\partial L}{\partial C^{[i]}}$ is $N_F^{[i]} \times N_{W}^{[i]}$, where $N_{W}^{[i]}= 2*N_{P}^{[i]}$.\\

\subsubsection{Convolutional Layer}
In this layer, backward propagation of the errors is carried out to calculate the fractional gradients for the kernel $D_{K^{[i]}}^{\alpha} L$  and bias $D_{b^{[i]}}^{\alpha} L$ matrices. The output gradient $\frac{\partial L}{\partial C^{[i]}}$ is used to calculate these gradients. Using Eqs. (\ref{convv3}) and (\ref{reg_l2}) we obtain

\begin{equation}
D_{K^{[i]}}^{\alpha} L = \dfrac{\partial L}{\partial S^{[i]}} D_{K^{[i]}}^{\alpha} S^{{[i]}} + \dfrac{\lambda}{2} D_{K^{[i]}}^{\alpha} (K^{[i]})^{2}
\label{c1xc}
\end{equation}

\begin{equation}
D_{b^{[i]}}^{\alpha} L = \dfrac{\partial L}{\partial S^{[i]}} D_{b^{[i]}}^{\alpha} S^{{[i]}}
\label{b1xc}
\end{equation}

Here, the gradient $\frac{\partial L}{\partial S^{[i]}}$ can be calculated using element-wise multiplication of $\frac{\partial L}{\partial C^{[i]}}$ and $\frac{\partial C^{[i]}}{\partial S^{[i]}}$ and is given as

\begin{equation}
    \frac{\partial L}{\partial S^{[i]}} = \frac{\partial L}{\partial C^{[i]}}\frac{\partial C^{[i]}}{\partial S^{[i]}}
\label{mkai}
\end{equation}

Using Eq. (\ref{relu1}), the value of $\frac{\partial C^{[i]}}{\partial S^{[i]}}$ can  be obtained as follows
\begin{equation}
\frac{\partial C^{[i]}}{\partial S^{[i]}} = 
\begin{dcases}
    1, & \text{if } S^{[i]} > 0\\
    0, & \text{if } S^{[i]}\leq 0
\end{dcases}
\end{equation}

Substituting the value of $\frac{\partial C^{[i]}}{\partial S^{[i]}}$ in Eq. ($41$), results in the following expression

\begin{equation}
\frac{\partial L}{\partial S^{[i]}} = 
\begin{dcases}
    \frac{\partial L}{\partial C^{[i]}}, & \text{if } S^{[i]} > 0\\
    0, & \text{if } S^{[i]}\leq 0
\end{dcases}
\label{kai}
\end{equation}

Using Eqs. (\ref{convv3}) and ($18$), the individual terms of Eq. (\ref{c1xc}) can be written in the following way
\begin{equation}
  D_{K^{[i]}}^{\alpha} S^{{[i]}} = I_{col}^{[i-1]} \left(\frac{(K^{[i]})^{1-\alpha}}{\Gamma (2-\alpha)}\right)
\label{c1}
\end{equation}
\begin{equation}
  \frac{\lambda}{2} D_{K^{[i]}}^{\alpha} {K^{[i]}}^2 = \lambda \frac{(K^{[i]})^{2-\alpha}}{\Gamma (3-\alpha)}
\label{c1}
\end{equation}

Substituting the above results in Eq. (\ref{c1xc}) the fractional kernel gradient is given as

\begin{equation}
\scalemath{0.8}{
D_{K^{[i]}}^{\alpha} L = 
\begin{dcases}
    \frac{\partial L}{\partial C^{[i]}} I_{col}^{[i-1]} \left(\frac{(K^{[i]})^{1-\alpha}}{\Gamma (2-\alpha)}\right) + \lambda \frac{(K^{[i]})^{2-\alpha}}{\Gamma (3-\alpha)}, & \text{if } S^{[i]} > 0\\
    \lambda \frac{(K^{[i]})^{2-\alpha}}{\Gamma (3-\alpha)}, & \text{if } S^{[i]}\leq 0
\end{dcases}
}
\end{equation}

Similarly, using Eqs. (\ref{mkai}) and (\ref{kai}) the fractional gradient with respect to the bias is given as
\begin{equation}
D_{b^{[i]}}^{\alpha} L = 
\begin{dcases}
    \frac{\partial L}{\partial C^{[i]}}\left(\frac{(b^{[i]})^{1-\alpha}}{\Gamma (2-\alpha)}\right), & \text{if } S^{[i]} > 0\\
    0, & \text{if } S^{[i]}\leq 0
\end{dcases}
\end{equation}

Next, for back-propagating the errors to the previous layers such as average pooling or up-sampling layer, the input gradient $\frac{\partial L}{\partial I^{[i-1]}}$  needs to be calculated. For this, first we need to calculate the gradient $\frac{\partial L}{\partial  I_{col}^{[i-1]}}$. It's value can be obtained using Eq. (\ref{convv3}) and is given as 

\begin{equation}
    \frac{\partial L}{\partial I_{col}^{[i-1]}} = \frac{\partial L}{\partial S^{[i]}} (K^{[i]})
\label{c1pp}
\end{equation}
with the dimension being $N_{C}^{[i-1]}\times F^{[i]}\times N_{W}^{[i]}$. Now, the value of  $\frac{\partial L}{\partial I^{[i-1]}}$ is obtained as follows
\begin{equation}
\frac{\partial L}{\partial I^{[i-1]}} = col2im \left(\frac{\partial L}{\partial I_{col}^{[i-1]}}\right)
\label{c1}
\end{equation}
with the dimension being $N_{C}^{[i-1]} \times N_{W}^{[i-1]}$. Here, the inverse transformation $col2im$ (Fig. \ref{col2im_fg}) is operated on  $\frac{\partial L}{\partial  I_{col}^{[i-1]}}$. Next, we use $\frac{\partial L}{\partial I^{[i-1]}}$ as $\frac{\partial L}{\partial U^{[i-1]}}$ or $\frac{\partial L}{\partial P^{[i-1]}}$ depending on whether the previous layer is an average pooling layer or the up-sampling layer. Once the gradients of all the trainable parameters are obtained, parameter update is carried out using gradient descent of learning rate $\eta$ as follows

\begin{equation}
   K^{[i]}\gets K^{[i]} - \eta D_{K^{[i]}}^{\alpha} L
\label{upd_1}
\end{equation}
\begin{equation}
   W\gets W - \eta D_{W}^{\alpha} L
\label{upd_3}
\end{equation}
\begin{equation}
   b^{[i]}\gets b^{[i]} - \eta D_{b^{[i]}}^{\alpha} L
\label{upd_2}
\end{equation}
\begin{equation}
   B\gets B - \eta D_{B}^{\alpha} L
\label{upd_4}
\end{equation}

\section{Compressed Architecture}
\label{cnn_cmp_sec}

\begin{figure}
\centering
 \includegraphics[width=0.5\textwidth,height=2.7in]{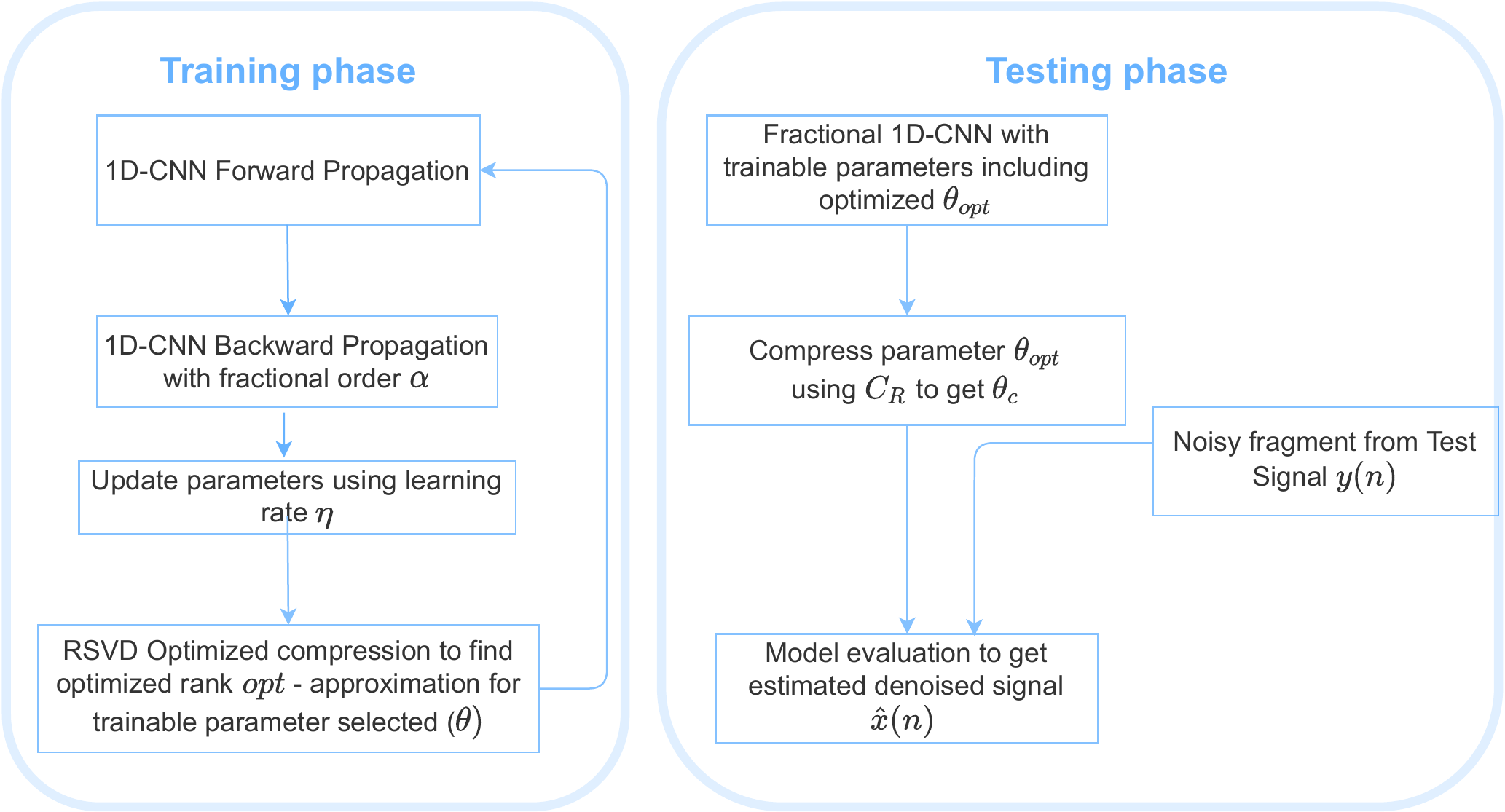}
\caption{Workflow of the proposed compressed fractional 1D-CNN}
\label{cnn_cmp_fg}
\end{figure}

The flowchart for the compressed version of the fractional based CNN architecture is shown in Fig. \ref{cnn_cmp_fg}. The training phase consists of forward and backward propagation of the fractional architecture discussed in Sec. \ref{sec_2}. This is followed by updating the trainable weights using fractional gradient descent. Next, the compression of the trained weights is carried out using RSVD function discussed in $algorithm$ 1. The inputs required for the calculation of RSVD are $\theta$ and $r$. Here, $r$ denotes the rank of the matrix whereas $\theta \in \{K^{[i]},W\}$ denotes the trainable set consisting of kernels $K^{[i]}$, which are used in convolution layers (CONV) and $W$ is the interconnection weight between the flattened and FC layers. The optimized rank ($opt$) is calculated as the rank at which $90\%$ of variance is covered for the singular vectors obtained using SVD decomposition given in Eq. ($15$). This is carried out using $check\_optimized\_rank$. The above description about the compression procedure carried out during the training process is summarized in $algorithm$ \ref{algo_1} and the steps are repeated till the convergence is achieved.

\begin{algorithm}
  \caption{RSVD compression for obtaining optimized Kernels and Weight matrices}
  \label{algo_1}
  \begin{algorithmic}[1]
  \State \textbf{procedure} $RSVD\_Opt\_Compression$($\theta$ , $r$):
 \State calculate [U\textsubscript{$r$} , S\textsubscript{$r$} , V\textsubscript{$r$}] = RSVD ($\theta$ , $r$) \algorithmiccomment from Eqs. $(11)$-$(16)$ 
 \State $opt \gets$  check\_optimized\_rank($S\textsubscript{r}$)
 \State $\theta_{opt} \gets U\textsubscript{opt}S\textsubscript{opt}V_{opt}^{T}$ 
 \State \textbf{return} $\theta_{opt}$
 \State \textbf{end procedure}
  \end{algorithmic}
\end{algorithm}

During the testing phase shown in Fig. \ref{cnn_cmp_fg}, the architecture has the optimized trainable parameter $\theta_{opt}$, obtained using $algorithm$ 1. Next, the compression of the $\theta_{opt}$ based on the compression rate ($C_{R}$) is performed using $algorithm$ 2 resulting in $\theta_{c}$ which is a rank $r$ approximation of the $\theta_{opt}$. Finally, the compressed parameter $\theta_{c}$ is used for denoising the EEG signals. The process is summarized in $algorithm$ \ref{algo_2}. Here, the compression rate $C_{R}$ is varied from $5\%$ to $95\%$ for our observations and this is done for various values of fractional order $\alpha$ ranging from $1$ to $1.5$.

\begin{algorithm}
  \caption{Evaluating denoising performance on testing dataset using compressed Kernels and Weight matrices}
  \label{algo_2}
  \begin{algorithmic}[1]
    \State \textbf{Input: } Trained parameter $\theta_{opt} \in \{K^{[i]},W\}$ 
    \State Initialize compression rate $C\textsubscript{R}$
     \State $r = (1-C\textsubscript{R}) * rank(\theta_{opt})$ 
 \State [U\textsubscript{$r$} , S\textsubscript{$r$} , V\textsubscript{$r$}] = RSVD ($\theta_{opt}$, $r$) \algorithmiccomment from Eqs.$(11)$-$(16)$ 
 \State $\theta_{c} \gets U\textsubscript{$r$}S\textsubscript{$r$}V_{r}^{T}$
    \State Estimated denoised signal $\hat x(n)$ using forward propagation \algorithmiccomment from Eqs. $(22)$-$(27)$ 
    \State \textbf{return} $\hat x(n)$
  \end{algorithmic}
\end{algorithm}

\begin{table*}
\captionsetup{justification=centering, labelsep=newline,font=footnotesize,labelfont=normalsize}
 \caption{COMPARISON OF PROPOSED ARCHITECTURE WITH EXISTING MA REMOVAL METHODS ON MENDELEY AND BONN(Z) DATABASE}
\label{my-label2}
  \centering
  \renewcommand{\arraystretch}{1.2}
    \begin{tabularx}{\textwidth}{|s|s|s|s|s|s|}
    
    \hline
    \textbf{Methods} & \textbf{Database}
    & \textbf{SNR} & \textbf{CC}  & \textbf{PRD} & \textbf{RMSE}
    \\
    \hline
    \multirow{2}{2cm}{EEMD-CCA\cite{b14}}   & Mendeley & 7.20  & 0.90 & 45.00 & 0.11\\ 
    & Bonn (Z) & 8.20  & 0.93 & 39.00 & 0.11 \\
    \hline
    \multirow{2}{2cm}{EEMD\cite{b13}}   & Mendeley & 7.46  & 0.90 & 43.61 & 0.11\\ 
    & Bonn (Z) & 8.44  & 0.93 & 38.01 & 0.10 \\
    \hline
    \multirow{2}{2cm}{EMD-CCA\cite{b14}}   & Mendeley & 4.46  & 0.81 & 59.56 & 0.15\\ 
    & Bonn (Z) & 5.45  & 0.85 & 53.81 & 0.15 \\
    \hline
    \multirow{2}{2cm}{Wavelet Denoising\cite{b11}}   & Mendeley & 7.10  & 0.90 & 45.51 & 0.11\\ 
    & Bonn (Z) & 7.70  & 0.92 & 41.52 & 0.11 \\
    \hline
    \multirow{2}{2cm}{EEMD-MCCA\cite{b15}}   & Mendeley & 7.26  & 0.90 & 44.78 & 0.11\\ 
    & Bonn (Z) & 8.26  & 0.93 & 38.94 & 0.11 \\
    \hline
    \multirow{2}{2cm}{VMD \cite{b16}}   & Mendeley & 8.00  & 0.92 & 41.46 & 0.10\\ 
    & Bonn (Z) & 9.00  & 0.95 & 36.13 & 0.09 \\
    \hline
    \multirow{2}{2cm}{Proposed Method}   & Mendeley & \textbf{8.68}  & \textbf{0.93} & \textbf{37.65} & \textbf{0.09}\\ 
    & Bonn (Z) & \textbf{10.15}  & \textbf{0.97} & \textbf{31.59} & \textbf{0.08} \\
    \bottomrule
  \end{tabularx}
\end{table*}

\begin{table*}
\captionsetup{justification=centering, labelsep=newline,font=footnotesize,labelfont=normalsize}
 \caption{PERFORMANCE OF PROPOSED MODEL ON MENDELEY AND BONN(Z) DATABASE}
\label{my-label1}
  \centering
  \renewcommand{\arraystretch}{1.2}
    \begin{tabularx}{\textwidth}{|s|s|s|s|s|s|}
    
    \hline
    \textbf{$\alpha$} & \textbf{Database}
    & \textbf{SNR} & \textbf{CC}  & \textbf{PRD} & \textbf{RMSE}
    \\
    \hline
    \multirow{2}{2cm}{1}   & Mendeley & 8.07  & 0.92 & 40.37 & 0.10\\ 
    & Bonn (Z) & 9.87  & 0.96 & 32.28 & 0.08 \\
    \hline
    \multirow{2}{2cm}{1.1}   & Mendeley & 8.41  & 0.93 & 38.80 & 0.10\\ 
    & Bonn (Z) & 9.85  & 0.96 & 32.46 & 0.08 \\
    \hline
    \multirow{2}{2cm}{1.2}   & Mendeley & \textbf{8.68}  & 0.93 & \textbf{37.65} & 0.09\\ 
    & Bonn (Z) & \textbf{10.15}  & 0.97 & \textbf{31.59} & 0.08 \\
    \hline
    \multirow{2}{2cm}{1.3}   & Mendeley & 8.32  & 0.92 & 39.14 & 0.10\\ 
    & Bonn (Z) & 9.16  & 0.95 & 35.25 & 0.09 \\
    \hline
    \multirow{2}{2cm}{1.4}   & Mendeley & 8.49  & 0.93 & 38.49 & 0.09\\ 
    & Bonn (Z) & 10.09  & 0.97 & 31.67 & 0.08 \\
    \hline
    \multirow{2}{2cm}{1.5}   & Mendeley & 8.50  & 0.92 & 38.48 & 0.10\\ 
    & Bonn (Z) & 8.85  & 0.94 & 36.62 & 0.10 \\
    \hline
    \multirow{2}{2cm}{1.6}   & Mendeley & 7.50  & 0.92 & 43.36 & 0.11\\ 
    & Bonn (Z) & 8.33  & 0.93 & 38.67 & 0.11 \\
    \bottomrule
  \end{tabularx}
\end{table*}

\section{Experiments and Evaluation}
In this section, several experiments are conducted to validate the efficiency of our proposed architecture. Firstly the standard datasets on which evaluations are conducted is presented, followed by experimental results including comparison with existing MA removal methods. Next, a detail study in which the performance of the architecture after compressing the kernel weights using low rank approximation is examined. All the experiments are performed on TESLA K80 GPU.

\begin{figure*}
\centering
 \includegraphics[width=\textwidth,height=2in]{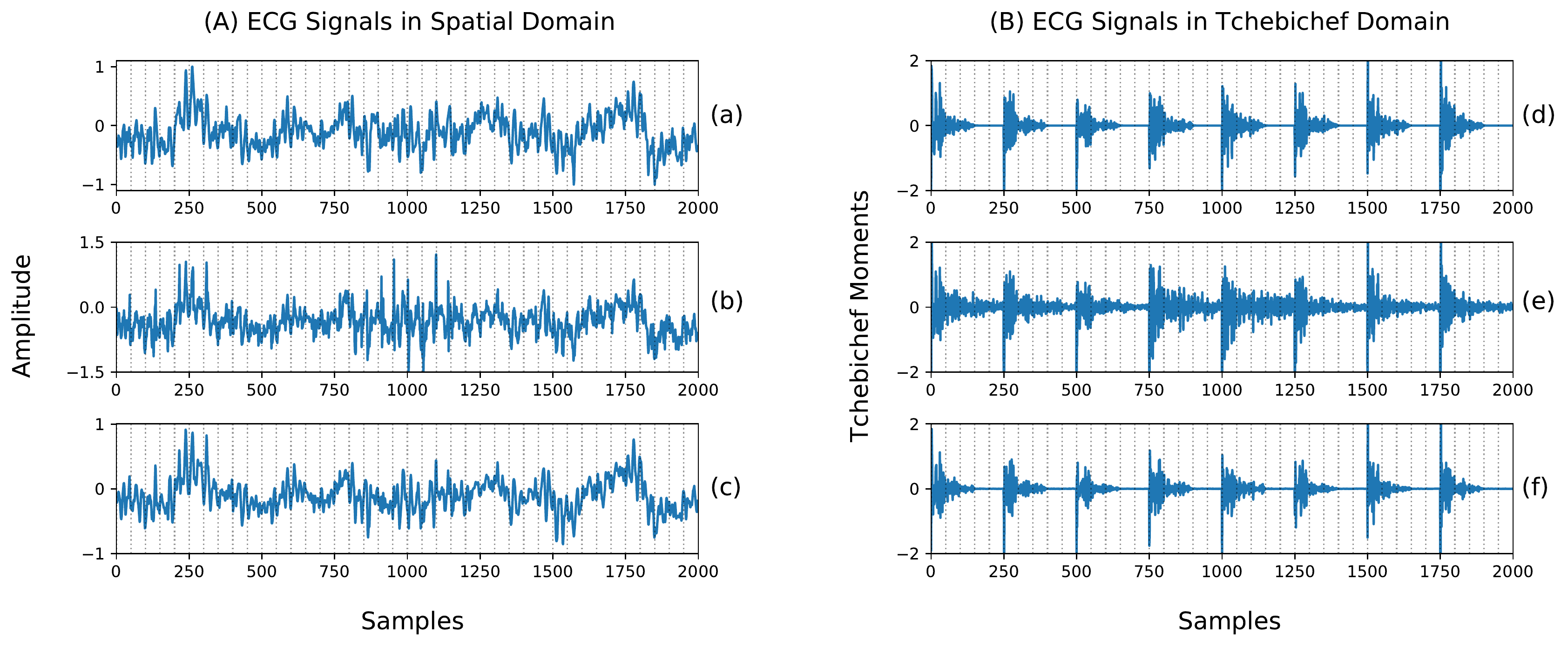}
\caption{EEG signals visualizations for Mendeley database. (A) Spatial domain: (a) Original  signal (b) Noisy signal (c) Denoised signal; (B) Tchebichef moment: (d)-(f) Corresponding signals in Tchebichef domain }
\label{mdl_fg}
\end{figure*}

\begin{figure*}
\centering
 \includegraphics[width=\textwidth,height=2in]{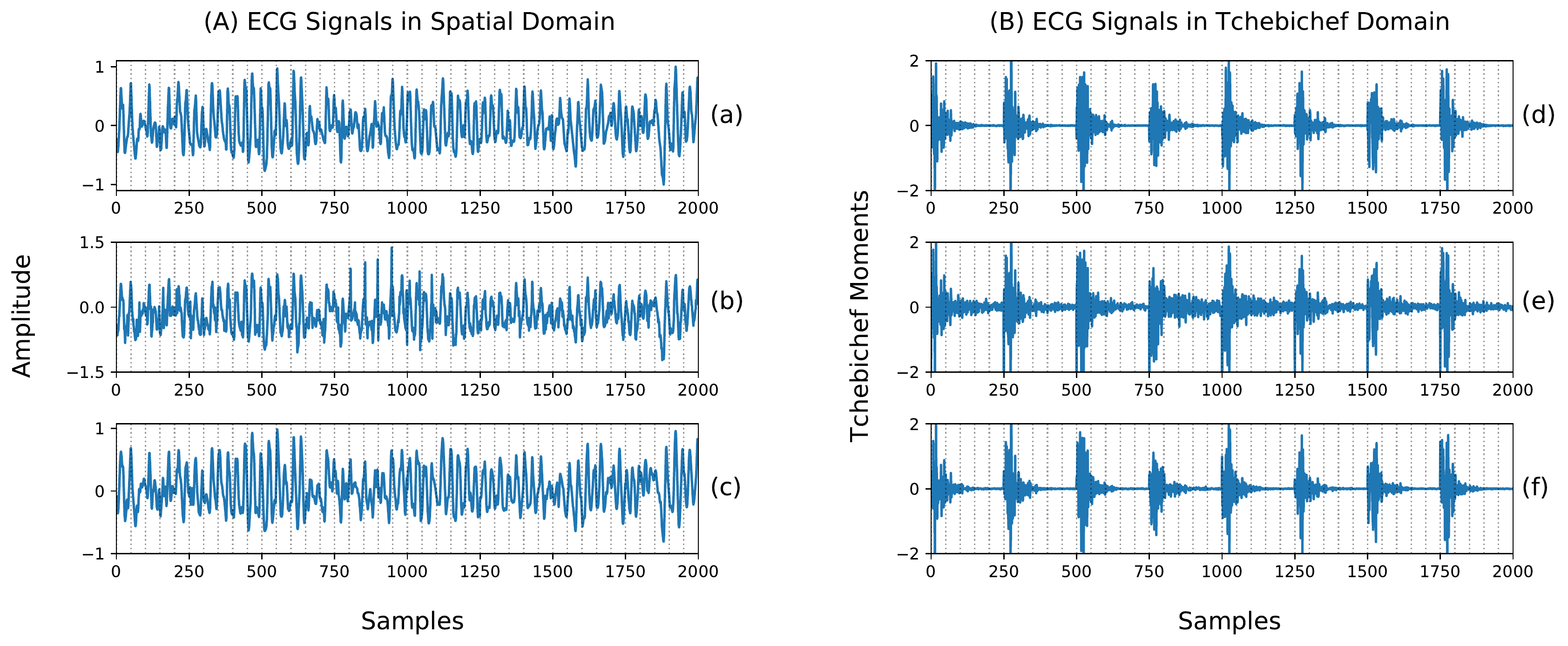}
\caption{ EEG signals visualizations for Bonn(Z) database. (A) Spatial domain: (a) Original  signal (b) Noisy signal (c) Denoised signal; (B) Tchebichef moment: (d)-(f) Corresponding signals in Tchebichef domain}
\label{bonn_fg}
\end{figure*}

\subsection{EEG Datasets}
Now, we evaluate the architecture on two publicly available databases, i.e., Mendeley and epileptic Bonn database. Mendeley database contains clean EEG recordings of 40 subjects, each having 19 channels and sampled at 200 Hz. The epileptic Bonn database contains five different sets of databases, each representing a particular subject. The subjects Z and O contains EEG recordings of five healthy subjects with eyes open and closed, respectively, subjects N and F contain inter-ictal recordings from seizure patients and S represents the seizure-EEG signals. These EEG signals are sampled at 173.61 Hz. For creating MA-contaminated EEG signals, we take the help from examples of electromyograms database, which has clean EMG signals recorded from healthy subjects, and patients with myopathy and neuropathy. The noisy signals are created by randomly mixing the clean EEG signals with EMG signals after re-sampling at 200 Hz.

\subsection{Data Preparation and Performance Metrics}

For Mendeley database, we took 1026 signals, each of 2000 samples, from which the training and testing data were created after splitting the data into $80\%$ training and $20\%$ testing set. 
Bonn database has five subjects each having 100 signals. We take those 100 signals and make a $80\%$-$20\%$ train-test split across all the subjects, i.e., $80$ signals are taken from each of the five subjects for training, while remaining $20$ for testing. Accordingly, the contaminated signals are generated by randomly mixing EMG signals with the original EEG ones. For comparison with existing MA removal methods, we take the subject 'Z' from Bonn database (represented by Bonn(Z)) for comparison while the whole testing set is taken from Mendeley database.

The next step involves creating fragments of $250$ samples from both the noisy and original EEG signals. The number of signals after taking each fragments is just $8$ times more, so it is not enough for training any deep neural network. To combat this, we perform data augmentation by randomly choosing a point from a particular signal to take $250$ fragments and repeat this process for few number of iterations such that finally, we have around $20000$ fragments for training. All these fragments are then transformed into orthogonal domain using TMs. After normalizing the fragments using $standardscalar$ function from $sklearn$\cite{b37} library, the resulting noisy Tchebichef vectors are fed as an input to the proposed architecture.

The evaluation of the proposed architecture is performed using different performance metrics such as signal-to-noise ratio ($SNR$), correlation coefficient ($CC$), percentage root mean square difference ($PRD$) and root mean square error ($RMSE$) calculated as 

\begin{equation}
   SNR = 10 * \log_{10} \frac{\sum\limits_{n=1}^{N}[x(n)]^{2}}{\sum\limits_{n=1}^{N}[x(n) - \hat x(n)]^{2}}
\label{snr1}
\end{equation}

\begin{equation}
   CC = \frac{\sum\limits_{n=1}^{N} (x(n)-M_x)(\hat x(n) - M_{\hat x})}{\sqrt{x(n)-M_x}\sqrt{\hat x(n) - M_{\hat x}}}
\label{cc1}
\end{equation}

\begin{equation}
   PRD = 100 * \sqrt{\frac{\sum\limits_{n=1}^{N}[x(n) - \hat x(n)]^{2}}{\sum\limits_{n=1}^{N}[x(n)]^{2}}}
\label{prd1}
\end{equation}

\begin{equation}
   RMSE = \sqrt{\frac{\sum\limits_{n=1}^{N}[x(n) - \hat x(n)]^{2}}{N}}
\label{rmse1}
\end{equation}
where $x(n)$ is the original signal, $\hat x(n)$ is the reconstructed signal, $M_x$ and $M_{\hat x}$ denote the mean of $x(n)$ and $\hat x(n)$ respectively,  and $N$ is the total number of samples.

To validate the effectiveness of the proposed architecture, comparison of the results are carried out with the existing MA removal methods, namely, Wavelet denoising \cite{b11}, EEMD \cite{b13}, EMD-CCA \cite{b14}, EEMD-CCA \cite{b14}, EEMD-MCCA \cite{b15} and VMD \cite{b16}

\begin{table*}
\captionsetup{justification=centering, labelsep=newline,font=footnotesize,labelfont=normalsize}
 \caption{LAYER WISE OPTIMIZED RANKS AFTER TRAINING THE PROPOSED ARCHITECTURE WITH RSVD COMPRESSION FOR MENDELEY AND BONN(Z) DATABASE}
\label{opt_rank}
  \centering
  \renewcommand{\arraystretch}{1.2}
    \begin{tabularx}{\textwidth}{sssssss}
    \toprule
    \toprule
    \multirow{2}{7cm}{\textbf{Database}}  & \multicolumn{2}{c}{\textbf{CONV2} } &
    \multicolumn{2}{c}{\textbf{CONV3} } &
    \multicolumn{2}{c}{\textbf{FC}}\\
    
    \cmidrule(lr){2-3}
    \cmidrule(lr){4-5}
    \cmidrule(lr){6-7}
    & \textbf{Original Rank} & \textbf{Optimized Rank} & \textbf{Original Rank} & \textbf{Optimized Rank} &
    \textbf{Original Rank} & \textbf{Optimized Rank}
    \\
    \cmidrule(lr){2-2}
    \cmidrule(lr){3-3}
    \cmidrule(lr){4-4}
    \cmidrule(lr){5-5}
    \cmidrule(lr){6-6}
    \cmidrule(lr){7-7}
    \textbf{Mendeley}   & 48 & 36  & 64 & 53 & 250  & 187\\ 
    \textbf{Bonn(Z)}  & 48 & 36  & 64 & 53 & 250 & 211\\ 
    \bottomrule
    \bottomrule
  \end{tabularx}
\end{table*}

\begin{figure*}
\centering
 \includegraphics[width=\textwidth,height=3in]{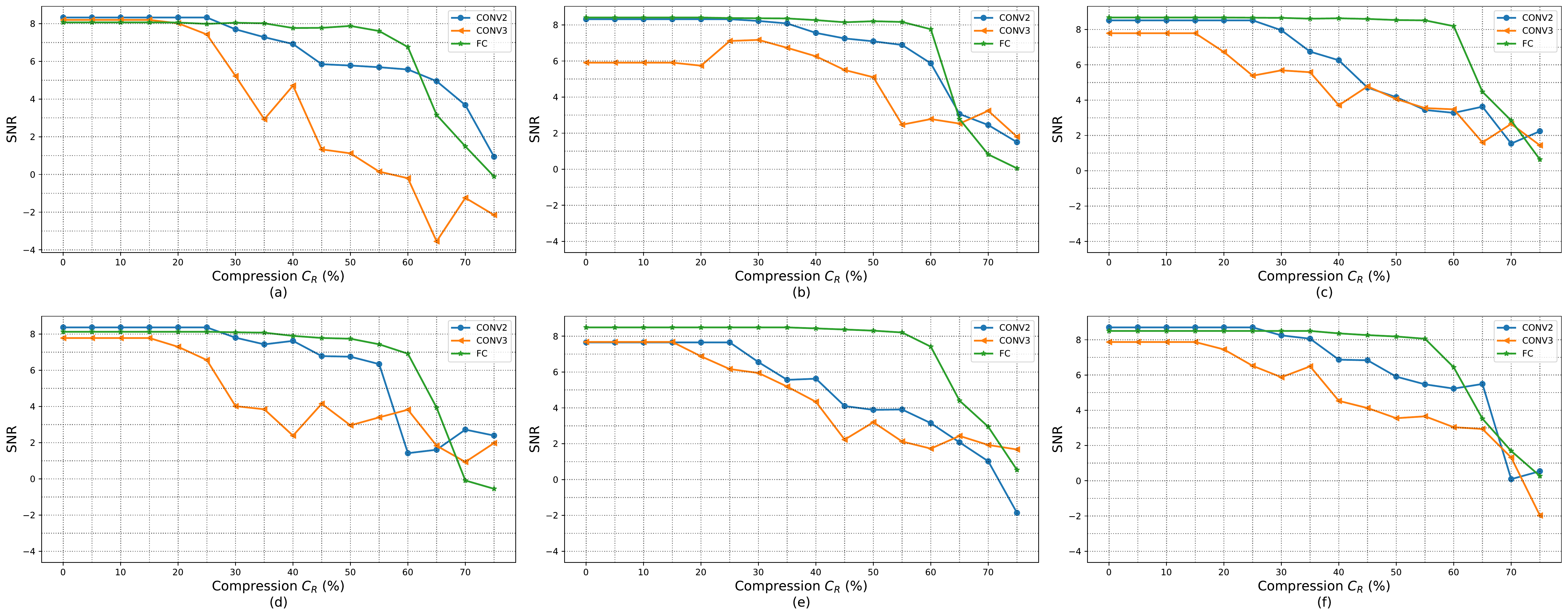}
\caption{SNR performance of the proposed architecture on Mendeley database after compression. Fractional order ($\alpha$) used are: (a) 1 , (b) 1.1, (c) 1.2, (d) 1.3, (e) 1.4, (f) 1.5}
\label{mdl_cmp}
\end{figure*}

\begin{figure*}
\centering
 \includegraphics[width=\textwidth,height=3in]{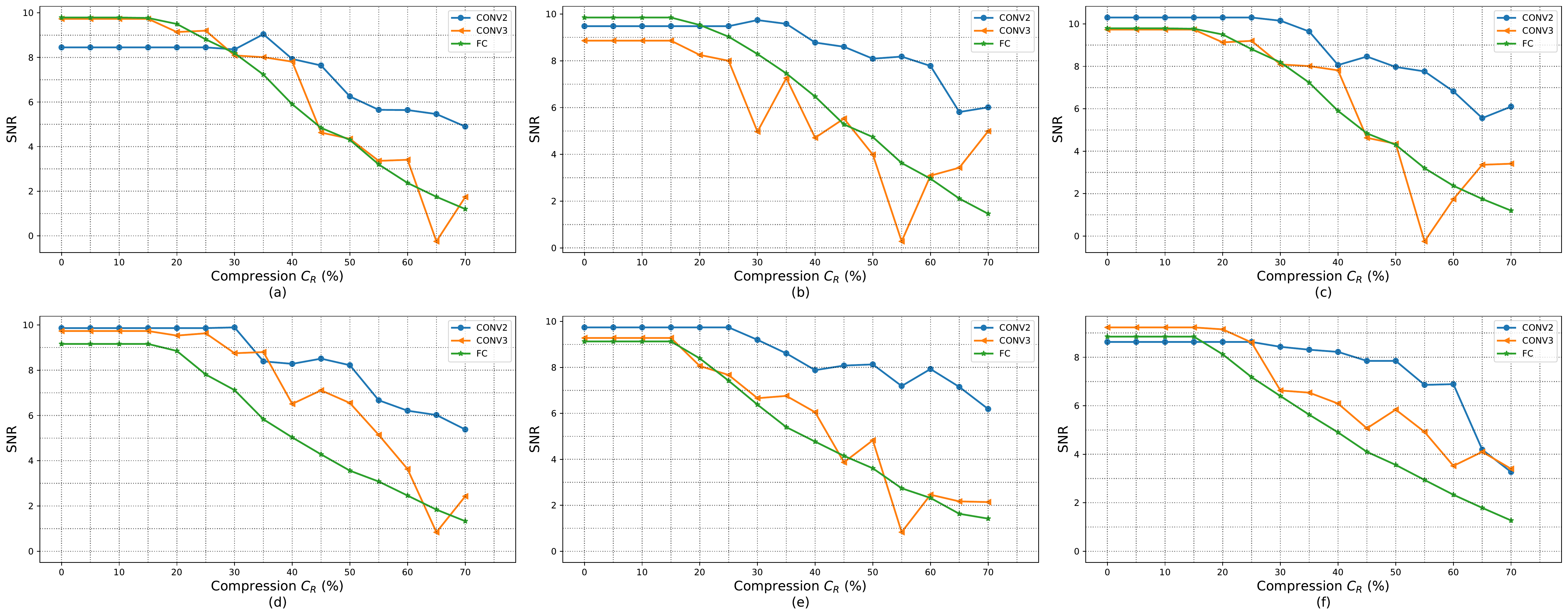}
\caption{SNR performance of the proposed architecture on Bonn(Z) database after compression. Fractional order ($\alpha$) used are: (a) 1 , (b) 1.1, (c) 1.2, (d) 1.3, (e) 1.4, (f) 1.5}
\label{bonn_cmp}
\end{figure*}

\subsection{Denoising Performance}
\begin{figure*}
\centering
 \includegraphics[width=\textwidth,height=3.5in]{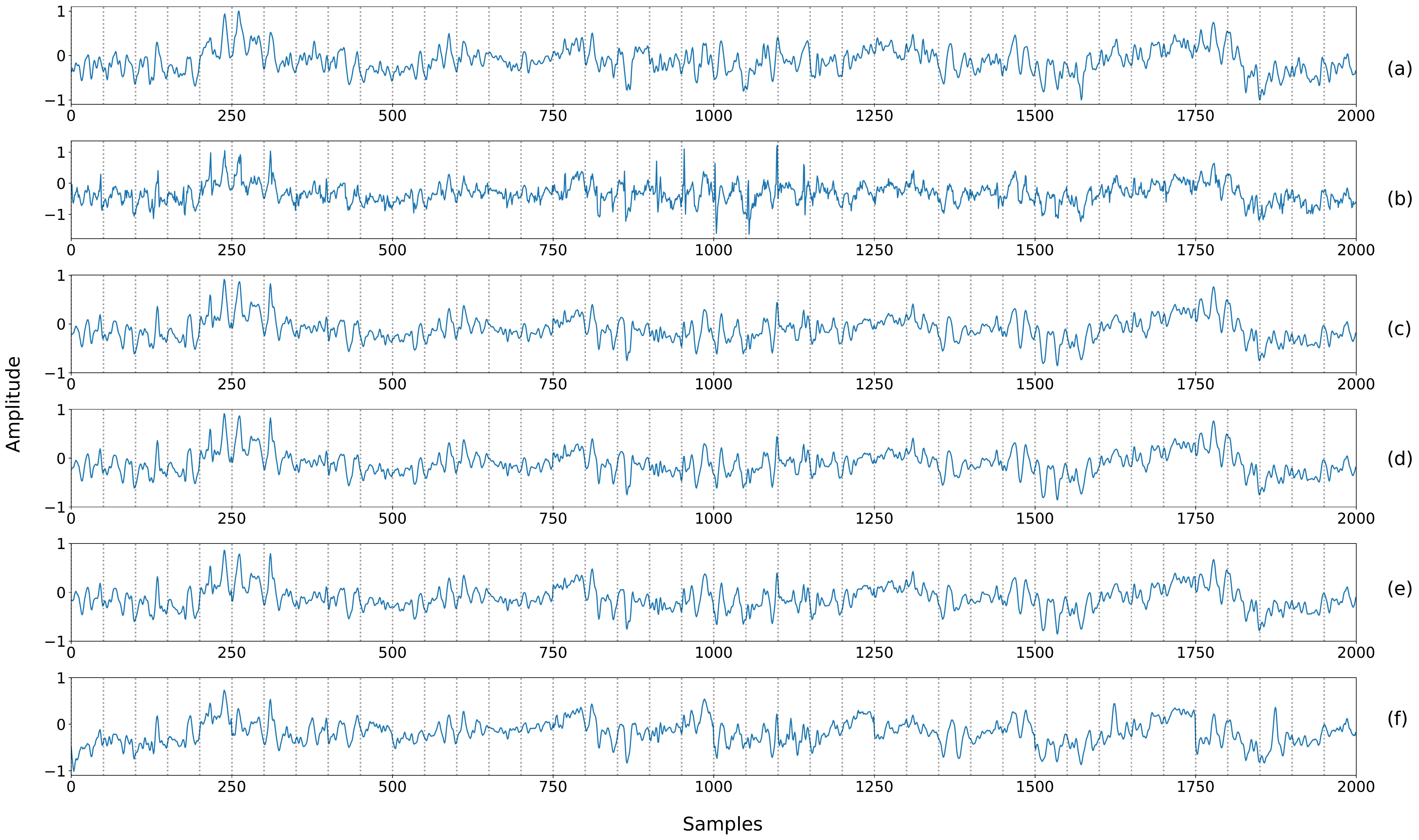}
\caption{EEG signals visualization under compression at various rates: (a) Original EEG signal , (b) Noisy EEG signal; Denoised signals: (c) without compression, (d) $25\%$ compression, (e)  $50\%$ compression, (f) $75\%$ compression}
\label{graph_cmp}
\end{figure*}

To validate the denoising performance of the architecture, a comparative analysis is carried out with the existing MA removal methods that have given good results on both the databases. It can be observed from Table \ref{my-label2}, that the performance metrics for the proposed fractional CNN architecture outperforms all of the existing methods. Our model gives an improvement of $8.5\%$ and $12.7\%$ in SNR values for Mendeley and Bonn(Z), respectively, when compared to the recently introduced variational mode decomposition (VMD) method.

The optimal hyper-parameters for the convolutional neural network were selected only after parameter tuning. Training and testing data loss was monitored after each epoch to check the condition of over-fitting. It was found that with batch size of $64$, learning rate ($\eta$) of $0.0005$, regularization parameter ($\lambda$) of $0.00001$ and training for $300$ epochs was found to be optimal for Mendeley database. For the Bonn database, training is performed using $200$ epochs while the other hyper-parameters remains the same.

Apart from the standard hyper-parameters listed in the aforementioned paragraph, the proposed architecture provides an extra hyper-parameter $\alpha$ that can be tuned to boost the denoising performance. The fractional order ($\alpha$) is used in calculating the weight gradients in back-propagation discussed in Sec. \ref{cai}.  Experiments are conducted, where $\alpha$ value is varied from $1$ to $1.6$ in steps of $0.1$ and the optimal one is taken for final evaluation. The performance metrics are represented in Table \ref{my-label1} after calculating the average over all the test data. It can be seen that for both the Mendeley and Bonn(Z) database, $\alpha = 1.2$ gives the best denoising results. Compared to the traditional integer order CNN with $\alpha = 1$, fractional CNN with $\alpha = 1.2$ gives around $7\%$ and $4\%$ performance boost in denoising for Mendeley and Bonn database respectively.

The original, noisy and the denoised EEG signals in spatial domain are visualized on left side of the Figs. (\ref{mdl_fg}) and (\ref{bonn_fg}) for Mendeley and Bonn(Z) database, respectively. The corresponding signals in Tchebichef domain are plotted on the right side. It can be observed from Figs. \ref{mdl_fg}-\ref{bonn_fg}(d)-(f)  that by transforming signals into orthogonal space using TMs exhibits sparse behaviour. These sparse signals are used as input feature vectors that helps in accelerating the training process as the architecture now requires fewer number of input coefficients to work upon.    

\subsection{Compression Analysis }

In this section, denoising results are presented when the fractional auto-encoder is trained with RSVD compression carried out on trainable weights (see Sect. \ref{cnn_cmp_sec}). The optimized rank for the weights matrix in FC layer and kernel matrix in CONV2 and CONV3 layer is selected such that $90\%$ of singular values can be retained. Table \ref{opt_rank} shows the original and the optimized rank for CONV2, CONV3 and FC layers  calculated using $algorithm$ \ref{algo_1}. It can be seen that for Bonn(Z) database, the optimized rank for FC layer is higher as compared to that of Mendeley database, which signifies that this layer will be more sensitive to compression for Bonn(Z) database.

The CONV2, CONV3 and FC layers are individually compressed using the optimized ranks obtained in Table \ref{opt_rank} and the performance of the model is evaluated using $algorithm$ \ref{algo_2}. Here, CONV1 layer is not compressed as it is directly interacting with the input. Fig. \ref{mdl_cmp} shows the effect of layer-wise compression on denoising performance evaluated in terms of SNR using the proposed model at various fractional orders. 
It can be observed that conventional CNN auto-encoder network shown in Fig. \ref{mdl_cmp}(a) with $\alpha = 1$ outperforms the existing state-of-the-art methods if CONV2 layer is compressed up-to $25\%$. However, as shown in Fig. \ref{mdl_cmp}(c), the proposed architecture at $\alpha = 1.2$ gives higher SNR compared to the conventional CNN even when the CONV2 layer is compressed by $30\%$. Likewise for FC layer compression, the architecture at $\alpha = 1.2$ gives better SNR values compared to other state of the art methods even at $60\%$ compression, while the conventional CNN outperforms other methods only upto $35\%$ compression. Moreover, the proposed architecture at $\alpha = 1.5$ (Fig. \ref{mdl_cmp}(f)) gives good performance under compression upto $55\%$ and $35\%$ of FC and CONV2, respectively. Similar analysis is  for the Bonn(Z) database and the results are shown in Fig. \ref{bonn_cmp}. 
It can be observed that the best SNR is obtained compared to the existing methods at $\alpha = 1.2$. This performance is guaranteed even when the CONV2, CONV3 and FC layers are compressed by $25\%$. For this database, it can be seen that $FC$ layer is more sensitive to compression. This is because the optimized rank (see Table \ref{opt_rank}) is $15\%$ of the original rank, as a result the compression at early stages leads to drop in the SNR performance. This is not in the case of Mendeley database as optimized rank is $25\%$ of the original rank. For further qualitative analysis, Fig. \ref{graph_cmp} shows the denoised signals produced by the architecture at $\alpha=1.2$ when CONV2 layer is subjected to various compression rates. It can be seen that even at $50\%$ compression the reconstructed signal resembles the original one at most of the points. Lastly, two things can be concluded from this study. Firstly, using hyper-parameter $\alpha$ can boost the SNR values of the EEG signals significantly. Secondly, by compressing the weights of the architecture does not degrade the SNR performance too much and its still better than the existing methods. Compressing the architecture provides an advantage that it consumes less memory space and is suitable for edge computing devices.

\section{Conclusion}

A fractional and compressed one-dimensional CNN auto-encoder, which uses orthogonal features in the form of Tchebichef moments has been proposed. The proposed method gives superior results in denoising MA contaminated signals when compared to existing MA removal methods with the best result observed at $\alpha=1.2$. Moreover, increasing the compression ratio ($C_{R}$) for the weights of the architecture, it beats the existing methods when evaluated using the performance metrics. Another important observation is that a compressed fractional architecture at $\alpha=1.2$ performs better than the conventional CNN auto-encoder without compression, i.e., with $60\%$ compression of FC layer for Mendeley database and $30\%$ compression of the CONV2 layer for the Bonn(Z) database. Compressing the architecture not only makes its occupies less memory foot-print but also delivers superior performance compared to other methods.
Qualitative analysis of the signals is presented at various stages of compression, which showed that the denoised signals are close to the original signals. Future work involves deployment of the compressed architecture on portable low energy devices.

\section{Appendix}

\subsection{Convergence Analysis}

A general fractional gradient descent update can be written as follows

\begin{equation}
    Z_{k} = Z_{k-1} - \eta D^{\alpha}_{Z_{k-1}} L
\end{equation}
\begin{equation}
    Z_{k+1} = Z_{k} - \eta D^{\alpha}_{Z_{k}} L
\end{equation}
where $k$ denotes the iteration. Using Eq. (\ref{frac_eq2}) the above equations can be re-written as

\begin{equation}
    Z_{k} = Z_{k-1} - \eta \frac{\partial L}{\partial Z_{k-1}} \frac{Z_{k-1}^{1-\alpha}}{\Gamma (2-\alpha)}
\end{equation}
\begin{equation}
    Z_{k+1} = Z_{k} - \eta \frac{\partial L}{\partial Z_{k}} \frac{Z_{k}^{1-\alpha}}{\Gamma (2-\alpha)}
\end{equation}

\textbf{Theorem 1: } \textit{If the fractional gradient descent method (59) is convergent, then it converges to a real extreme point $z^{*}$}

We can prove it by contradiction. Let's assume $Z_{k}$ converges to a different point $Z\neq z^{*}$, thus 
\begin{equation}
    \lim_{k\to\infty} \abs{Z_{k}-Z} = 0 
\end{equation}
Let $\epsilon$ be any sufficiently small value. Then for any $\epsilon$, there exists a sufficiently large value $E\in \mathbb{N}$ such that $\abs{Z_{k-1} - Z} <  \epsilon < \abs{z^{*} - Z}$ for any $k-1 > E$ . On basis of assumption, $\abs{Z_{k} - Z} <  \epsilon$. Then  the following equation must hold,
\begin{equation}
    \inf_{k>E} \frac{\partial L}{\partial Z_{k}} > 0
\end{equation}
\begin{equation}
    Z_{k+1} - Z_{k} =\abs{\eta \frac{\partial L}{\partial Z_{k}} \frac{Z_{k}^{1-\alpha}}{\Gamma (2-\alpha)}}
\end{equation}

\begin{equation}
   \abs{ Z_{k+1} - Z_{k} } \geqslant d \abs{Z_{k}^{1-\alpha}}
   \label{kap3}
\end{equation}
where $d = \eta \frac{1}{\Gamma (2-\alpha)} \inf_{k>E} \frac{\partial L}{\partial Z_{k}}$.

Assuming there is a $\epsilon$ such that $2\epsilon < d^{1/\alpha}$, then
\begin{equation}
    \abs{Z_{k} - Z_{k-1}} < \abs{Z_{k} - Z} + \abs{Z_{k-1}-Z} < 2\epsilon < d^{1/\alpha}
    \label{kap1}
\end{equation}

From Eq. (\ref{kap1}), we can write
\begin{equation}
    \abs{Z_{k} - Z_{k-1}}^{\alpha} < d
    \label{kap2}
\end{equation}

Combining Eqs. \ref{kap3} and \ref{kap2} we obtain
\begin{equation}
   \abs{Z_{k+1} - Z_{k}} > \abs{Z_{k} - Z_{k-1}}
\end{equation}

Above equation implies that $Z_{k}$ is not convergent. It contradicts the assumption that $Z_{k}$ is convergent to $Z$ and thus the proof is completed.

\textbf{Theorem 2: } \textit{The FC layer updated by the fractional gradient descent performed using Eqs. (\ref{upd_3}) and (\ref{upd_4}) are convergent to real extreme point}

From Eq. (\ref{upd_3}), update equation for each element of weight matrix $W$ can be given by
\begin{equation}
   W_{ij(k+1)}\gets W_{ij(k)} - \eta \frac{\partial L}{\partial W_{ij(k)}} \frac{W_{ij(k)}^{1-\alpha}}{\Gamma (2-\alpha)} 
   \label{cap1}
\end{equation}
\begin{equation}
   B_{i(k+1)}\gets B_{i(k)} - \eta \frac{\partial L}{\partial B_{i(k)}} \frac{B_{i(k)}^{1-\alpha}}{\Gamma (2-\alpha)} 
   \label{cap2}
\end{equation}
The remaining part is similar to the proof of Theorem 1.

Let's assume $W_{ij(k)}$ converges to a different point $W_{ij}^{'}\neq W_{ij}^{*}$, where $W_{ij}^{*}$ is the real extreme point. Thus 
\begin{equation}
    \lim_{k\to\infty} \abs{W_{ij(k)}-W_{ij}^{'}} = 0 
\end{equation}
Let $\epsilon$ be any sufficiently small value. Then for any $\epsilon$, there exists a sufficiently large value $E\in \mathbb{N}$ such that $\abs{W_{ij(k-1)} - W_{ij}^{'}} <  \epsilon < \abs{W_{ij}^{*} - W_{ij}^{'}}$ for any $k-1 > E$ . On basis of assumption, $\abs{W_{ij(k-1)} - W_{ij}^{'}} <  \epsilon$. Then  the following equation must hold,
\begin{equation}
    \inf_{k>E} \frac{\partial L}{\partial W_{ij(k)}} > 0
\end{equation}
\begin{equation}
    W_{ij(k+1)} - W_{ij(k)} =\abs{\eta \frac{\partial L}{\partial W_{ij(k)}} \frac{W_{ij(k)}^{1-\alpha}}{\Gamma (2-\alpha)}}
\end{equation}

\begin{equation}
   \abs{W_{ij(k+1)} - W_{ij(k)} } \geqslant d \abs{W_{ij(k)}^{1-\alpha}}
\end{equation}
where $d = \eta \frac{1}{\Gamma (2-\alpha)} \inf_{k>E} \frac{\partial L}{\partial W_{ij(k)}}$.

Assuming there is a $\epsilon$ such that $2\epsilon < d^{1/\alpha}$, then
\begin{equation}
    \abs{W_{ij(k)} - W_{ij(k-1)}} < \abs{W_{ij(k)} - W_{ij}^{'}} + \abs{W_{ij(k-1)}-W_{ij}^{'}} < 2\epsilon < d^{1/\alpha}
\end{equation}

From Eq. $75$, we can write
\begin{equation}
    \abs{W_{ij(k)} - W_{ij(k-1)}}^{\alpha} < d
\end{equation}

Combining Eq. $74$ and $76$,
\begin{equation}
   \abs{W_{ij(k+1)} - W_{ij(k)}} > \abs{W_{ij(k)} - W_{ij(k-1)}}
\end{equation}

Above equation implies that $W_{ij(k)}$ is not convergent. Similarly, same result can be obtained for bias $B_{i(k)}$. It contradicts the assumption that $W_{ij(k)}$ is convergent to $W_{ij}^{'}$ and thus the proof is completed.

\textbf{Theorem 3: } \textit{The CONV layers updated by the fractional gradient descent method Eqs. (\ref{upd_1}) and (\ref{upd_2}) are convergent to real extreme point}

The update equations given in Eqs. (\ref{upd_1}) and (\ref{upd_2}) can be written in the form as mentioned in  Eqs. (\ref{cap1})-(\ref{cap2}). After that the proof resembles Theorem 2.

\bibliographystyle{unsrt}
\bibliography{references.bib}

\begin{thebibliography}{10}

\bibitem{b1}
F.~C. {Morabito}, D.~{Labate}, A.~{Bramanti}, F.~L. {Foresta}, G.~{Morabito},
  I.~{Palamara}, and H.~H. {Szu}.
\newblock ``{Enhanced Compressibility of EEG Signal in Alzheimer's Disease
  Patients}''.
\newblock {\em IEEE Sensors Journal}, vol. 13, no. 9, pp. 3255--3262, 2013.

\bibitem{b2}
Rishi Sharma, Piyush Varshney, Ram Pachori, and Santosh Vishvakarma.
\newblock ``{Automated System for Epileptic EEG Detection Using Iterative
  Filtering}''.
\newblock {\em IEEE Sensors Letters}, vol. 2, pp. 1--4, Nov 2018.

\bibitem{b3}
Jiang X, Bian GB, and Tian Z.
\newblock ``{Removal of Artifacts from EEG Signals: A Review}''.
\newblock {\em Sensors}, vol. 19, no. 5, 2019.

\bibitem{b4}
X.~{Chen}, X.~{Xu}, A.~{Liu}, S.~{Lee}, X.~{Chen}, X.~{Zhang}, M.~J. {McKeown},
  and Z.~J. {Wang}.
\newblock ``{Removal of Muscle Artifacts From the EEG: A Review and
  Recommendations}''.
\newblock {\em IEEE Sensors Journal}, vol. 19, no. 14, pp. 5353--5368, 2019.

\bibitem{b5}
N.~V. {Thakor} and Y.~. {Zhu}.
\newblock ``{Applications of adaptive filtering to ECG analysis: noise
  cancellation and arrhythmia detection}''.
\newblock {\em IEEE Transactions on Biomedical Engineering}, vol. 38, no. 8,
  pp. 785--794, 1991.

\bibitem{b6}
T.~.~M. {Slonim}, M.~A. {Slonim}, and E.~A. {Ovsyscher}.
\newblock ``{The use of simple FIR filters for filtering of ECG signals and a
  new method for post-filter signal reconstruction}''.
\newblock In {\em Proceedings of Computers in Cardiology Conference}, pages
  871--873, 1993.

\bibitem{b7}
V.~X. {Afonso}, W.~J. {Tompkins}, T.~Q. {Nguyen}, S.~{Trautmann}, and S.~{Luo}.
\newblock ``{Filter bank-based processing of the stress ECG}''.
\newblock In {\em Proceedings of 17th International Conference of the
  Engineering in Medicine and Biology Society}, volume~2, pages 887--888, 1995.

\bibitem{b8}
Noor Al-Qazzaz.
\newblock ``{Noise Removal of ECG Signal Using Recursive Least Square
  Algorithms}''.
\newblock {\em Al-Khwarizmi Eng. J.}, vol. 7, no. 1, pp. 13--21, Jan 2011.

\bibitem{b9}
Md~Islam, G~M~Sabil Sajjad, Hamidur Rahman, Ajoy Dey, Md~Biswas, and A.~Hoque.
\newblock ``{Performance Comparison of Modified LMS and RLS Algorithms in
  Denoising of ECG Signals}''.
\newblock {\em International Journal of Engineering and Technology}, vol. 2,
  pp. 466--468, Mar 2012.

\bibitem{b10}
O.~{Sayadi} and M.~B. {Shamsollahi}.
\newblock ``{ECG Denoising and Compression Using a Modified Extended Kalman
  Filter Structure}''.
\newblock {\em IEEE Transactions on Biomedical Engineering}, vol. 55, no. 9,
  pp. 2240--2248, 2008.

\bibitem{b11}
F.~La~Foresta N.~Mammone and F.~C. Morabito.
\newblock ``{Automatic artifact rejection from multichannel scalp EEG by
  wavelet ICA}''.
\newblock {\em IEEE Sensors Journal}, vol. 12, no. 3, pp. 533--542, 2012.

\bibitem{b12}
N.~E. Huang, Z.~Shen, and S.~R.~Long et~al.
\newblock ``{The empirical mode decomposition and the Hilbert spectrum for
  nonlinear and non-stationary time}''.
\newblock In {\em Proceedings of the Royal Society of London A}, volume vol.
  454, pp. 903--995, 1998.

\bibitem{b13}
Z.~Wu and N.~E. Huang.
\newblock ``{Ensemble empirical mode decomposition: A noise-assisted data
  analysis method}''.
\newblock {\em Advances in Adaptive Data Analysis}, vol. 1, no. 1, pp. 1--41,
  2009.

\bibitem{b14}
K.~T. {Sweeney}, S.~F. {McLoone}, and T.~E. {Ward}.
\newblock ``{The Use of Ensemble Empirical Mode Decomposition With Canonical
  Correlation Analysis as a Novel Artifact Removal Technique}''.
\newblock {\em IEEE Transactions on Biomedical Engineering}, vol. 60, no. 1,
  pp. 97--105, 2013.

\bibitem{b15}
Hu~Peng Xun~Chen, Chen~He.
\newblock ``{Removal of Muscle Artifacts from Single-Channel EEG Based on
  Ensemble Empirical Mode Decomposition and Multiset Canonical Correlation
  Analysis}''.
\newblock {\em Journal of Applied Mathematics}, vol. 2014, 2014.

\bibitem{b16}
Upadhayay~MD Saini~M, Satija~U.
\newblock ``{Effective automated method for detection and suppression of muscle
  artefacts from single-channel EEG signal}''.
\newblock {\em Journal of Applied Mathematics}, vol. 7, pp. 35--40, 2020.

\bibitem{b17}
Pascal Vincent, Hugo Larochelle, Y.~Bengio, and Pierre-Antoine Manzagol.
\newblock ``{Extracting and composing robust features with denoising
  autoencoders}''.
\newblock In {\em Proceedings of the 25th International Conference on Machine
  Learning}, pages 1096--1103, Jan 2008.

\bibitem{b18}
D.~{Yu} and L.~{Deng}.
\newblock ``{Deep Learning and Its Applications to Signal and Information
  Processing [Exploratory DSP]}''.
\newblock {\em IEEE Signal Processing Magazine}, vol. 28, no. 1, pp. 145--154,
  2011.

\bibitem{b19}
Peng Xiong, Hongrui Wang, Ming Liu, and Xiuling Liu.
\newblock ``{Denoising Autoencoder for Eletrocardiogram Signal Enhancement}''.
\newblock {\em Journal of Medical Imaging and Health Informatics}, vol. 5, pp.
  1804--1810, Dec 2015.

\bibitem{b20}
Peng Xiong, Hongrui Wang, Ming Liu, Feng Lin, Zengguang Hou, and Xiuling Liu.
\newblock ``{A stacked contractive denoising auto-encoder for {ECG} signal
  denoising}''.
\newblock {\em Physiological Measurement}, vol. 37, no. 12, pp. 2214--2230, Nov
  2016.

\bibitem{b21}
Song Han, Jeff Pool, John Tran, and William Dally.
\newblock ``{Learning both Weights and Connections for Efficient Neural
  Network}''.
\newblock In {\em Advances in Neural Information Processing Systems}, volume
  vol. 28. Curran Associates, Inc., 2015.

\bibitem{b22}
Max Jaderberg, Andrea Vedaldi, and Andrew Zisserman.
\newblock ``{Speeding up Convolutional Neural Networks with Low Rank
  Expansions}''.
\newblock In {\em Proceedings of the British Machine Vision Conference}. BMVA
  Press, 2014.

\bibitem{b23}
Xiyu Yu, Tongliang Liu, Xinchao Wang, and Dacheng Tao.
\newblock ``{On Compressing Deep Models by Low Rank and Sparse
  Decomposition}''.
\newblock In {\em Proceedings of the IEEE Conference on Computer Vision and
  Pattern Recognition (CVPR)}, July 2017.

\bibitem{b34}
N.~Benjamin Erichson, Sergey Voronin, Steven~L. Brunton, and J.~Nathan Kutz.
\newblock ``{Randomized Matrix Decompositions Using R}''.
\newblock {\em Journal of Statistical Software}, vol. 89, no. 11, 2019.

\bibitem{b35}
E.~{Clark}, S.~L. {Brunton}, and J.~N. {Kutz}.
\newblock ``{Multi-Fidelity Sensor Selection: Greedy Algorithms to Place Cheap
  and Expensive Sensors With Cost Constraints}''.
\newblock {\em IEEE Sensors Journal}, vol. 21, no. 1, pp. 600-611, 2021.

\bibitem{b24}
X.~{Zou}, X.~{Xu}, C.~{Qing}, and X.~{Xing}.
\newblock ``{High speed deep networks based on Discrete Cosine
  Transformation}''.
\newblock In {\em 2014 IEEE International Conference on Image Processing
  (ICIP)}, pages 5921--5925, 2014.

\bibitem{b25}
J.~{Zhao}, R.~{Xiong}, J.~{Xu}, F.~{Wu}, and T.~{Huang}.
\newblock ``{Learning a Deep Convolutional Network for Subband Image
  Denoising}''.
\newblock In {\em 2019 IEEE International Conference on Multimedia and Expo
  (ICME)}, pages 1420--1425, 2019.

\bibitem{b27}
Haiyong Wu and Senlin Yan.
\newblock ``{Computing invariants of Tchebichef moments for shape based image
  retrieval}''.
\newblock {\em Neurocomputing}, vol. 215, pp. 110--117, 2016.
\newblock SI: Stereo Data.

\bibitem{b28}
Huaining Cheng and Soon~M. Chung.
\newblock ``{Orthogonal moment-based descriptors for pose shape query on 3D
  point cloud patches}''.
\newblock {\em Pattern Recognition}, vol. 52, pp. 397--409, 2016.

\bibitem{b29}
A.~{Kumar}, M.~{Omair Ahmad}, and M.~N.~S. {Swamy}.
\newblock ``{Tchebichef and Adaptive Steerable-Based Total Variation Model for
  Image Denoising}''.
\newblock {\em IEEE Transactions on Image Processing}, vol. 28, no. 6, pp.
  2921--2935, 2019.

\bibitem{b30}
J.~{Yu}, L.~{Tan}, S.~{Zhou}, L.~{Wang}, and M.~A. {Siddique}.
\newblock ``{Image Denoising Algorithm Based on Entropy and Adaptive Fractional
  Order Calculus Operator}''.
\newblock {\em IEEE Access}, vol. 5, pp. 12275-12285, 2017.

\bibitem{b31}
Bo~Li and Wei Xie.
\newblock ``{Image denoising and enhancement based on adaptive fractional
  calculus of small probability strategy}''.
\newblock {\em Neurocomputing}, vol. 175, pp. 704--714, 2016.

\bibitem{b32}
Y.~{Pu}, J.~{Zhou}, and X.~{Yuan}.
\newblock ``{Fractional Differential Mask: A Fractional Differential-Based
  Approach for Multiscale Texture Enhancement}''.
\newblock {\em IEEE Transactions on Image Processing}, vol. 19, no. 2, pp.
  491--511, 2010.

\bibitem{b33}
Jian Wang, Yanqing Wen, Yida Gou, Zhenyun Ye, and Hua Chen.
\newblock ``{Fractional-order gradient descent learning of BP neural networks
  with Caputo derivative}''.
\newblock {\em Neural Networks}, vol. 89, pp. 19--30, 2017.

\bibitem{b26}
R.~{Mukundan}.
\newblock ``{Some computational aspects of discrete orthonormal moments}''.
\newblock {\em IEEE Transactions on Image Processing}, vol. 13, no. 8, pp.
  1055--1059, 2004.

\bibitem{b36}
Yi~Zhang Chunhui~Bao, Yifei~Pu.
\newblock ``{Fractional-Order Deep Backpropagation Neural Network}''.
\newblock {\em Computational Intelligence and Neuroscience}, vol. 2018, 2018.

\bibitem{b37}
F.~{Pedregosa}, G.~{Varoquaux}, A.~{Gramfort}, V.~{Michel}, B.~{Thirion},
  O.~{Grisel}, M.~{Blondel}, P.~{Prettenhofer}, R.~{Weis}, V.~{Dubourg},
  J.~{Vanderplas}, A.~{Passos}, D.~{Cournapeau}, M.~Brucher, M.~{Perrot}, and
  E.~{Duchesnay}.
\newblock ``{Scikit-learn: Machine learning in python}''.
\newblock {\em Journal of Machine Learning Research}, vol. 12, pp. 2825-2830,
  2018.

\end{thebibliography}

\end{document}